\documentclass[10pt,twocolumn,letterpaper]{article}

\usepackage[pagenumbers]{cvpr} 

\usepackage{graphicx}
\usepackage{amsmath}
\usepackage{amssymb}
\usepackage{booktabs}
\usepackage{cuted}
\usepackage{capt-of}
\usepackage{graphicx}
\usepackage{comment}
\usepackage{multirow}
\usepackage[all]{nowidow}

\usepackage[pagebackref,breaklinks,colorlinks]{hyperref}
\usepackage[accsupp]{axessibility}  

\usepackage[capitalize]{cleveref}
\Crefname{section}{Sec.}{Secs.}
\crefname{section}{Section}{Sections}
\crefname{table}{Table}{Tables}
\Crefname{table}{Tab.}{Tabs.}
\crefname{figure}{Figure}{Figures}

\def\cvprPaperID{3} 
\def\confName{CVPR}
\def\confYear{2023}

\newcommand{\method}[1]{Our Method}

\begin{document}

\title{CAMM: Building Category-Agnostic and Animatable 3D Models from Monocular Videos}

\author{Tianshu Kuai \; Akash Karthikeyan \; Yash Kant \; Ashkan Mirzaei \; Igor Gilitschenski\\
[4pt]
University of Toronto \\
}
\maketitle

\begin{strip}\centering
\vspace{-35pt}
\includegraphics[width=\linewidth, trim={0 0cm 0 0cm},clip]{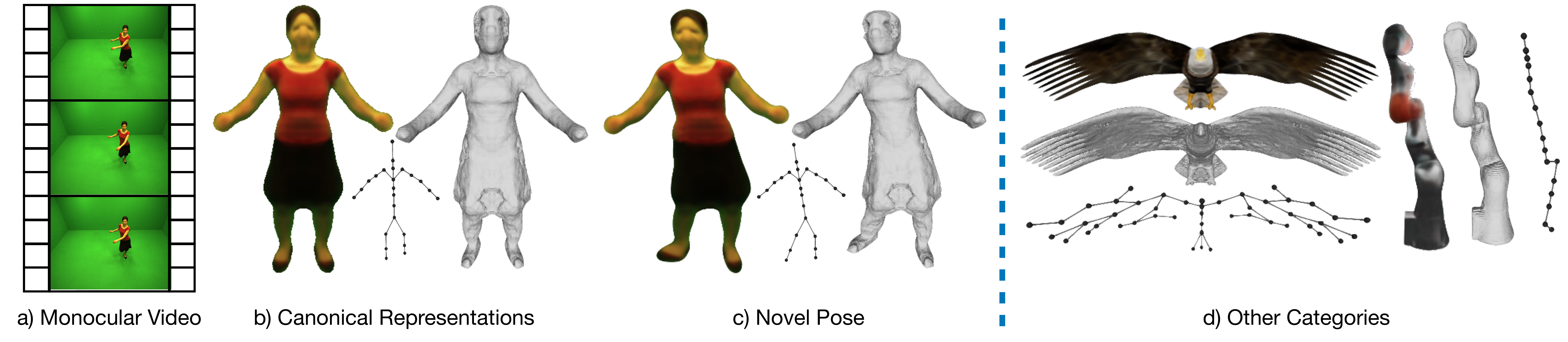}
\captionof{figure}{Given monocular videos of an articulated object, our method builds its canonical representation, including 3D shape, appearance, and a corresponding animatable 3D kinematic chain for direct pose manipulations. Our approach does not rely on any information on the object's shape and underlying structure. In c) we show example of re-posing the human in b) to novel pose. We show the learned canonical representations of other object categories in d).}
\label{fig:teaser}
\end{strip}

\begin{abstract}
Animating an object in 3D often requires an articulated structure, e.g. a kinematic chain or skeleton of the manipulated object with proper skinning weights, to obtain smooth movements and surface deformations. However, existing models that allow direct pose manipulations are either limited to specific object categories or built with specialized equipment. To reduce the work needed for creating animatable 3D models, we propose a novel reconstruction method that learns an animatable kinematic chain for any articulated object. Our method operates on monocular videos without prior knowledge of the object's shape or underlying structure. Our approach is on par with state-of-the-art 3D surface reconstruction methods on various articulated object categories while enabling direct pose manipulations by re-posing the learned kinematic chain. Our project page: \href{https://camm3d.github.io/}{https://camm3d.github.io/}. 
\end{abstract}
\section{Introduction}
\label{sec:intro}

Building 3D representations of real-world objects from images has long been studied in 3D Computer Vision. Realistic 3D models allow us to synthesize new images of objects from arbitrary viewpoints and to animate these objects for applications such as virtual and augmented reality. Prior works on building 3D representations for static scenes~\cite{mildenhall2020nerf,barron2021mip,choi2019extreme,lin2021barf,ruckert2022adop,yu2021plenoctrees,zhou2018stereo,fridovich2022plenoxels,jeong2021self,martin2021nerf,meng2021gnerf,self-sup-appearance} assume the objects in the scene to be rigid to obtain accurate correspondences, which cannot be generalized to deformable objects or scenes with drastic motions. Recent works~\cite{li2021neural,pumarola2021d,tretschk2021non,wang2021neural,kong2019deep,kumar2020non,sidhu2020neural,park2021nerfies,yang2021lasr,yang2021viser,banmo} tackle the problem of reconstructing dynamic scenes and 3D deformable objects, and have achieved impressive performance on novel view synthesis. However, these methods can only render objects with the body poses that exist in the given training data because they ``mimic'' the movements and behaviours of the objects. Therefore, their representations of 3D objects are not applicable for direct pose manipulations or animations. 

To fulfill such needs, tremendous efforts have been made to build parametric shape models for humans~\cite{loper2015smpl,pavlakos2019expressive,xiang2019monocular,xu2020ghum} and quadruped animals~\cite{zuffi20173d,zuffi2018lions}. These shape models are often built via capturing large amounts of data using specialized scanners and sensors. They serve as base templates for methods that focus on reconstructing and animating 3D humans and animals~\cite{vo2020spatiotemporal,peng2021neural,deng2020nasa,noguchi2021neural,weng2022humannerf,peng2021animatable,liu2021neural,kundu2020appearance,wang2022arah}. Although these template-based models can offer high-fidelity reconstruction and animation results, they can only be applied to limited object categories. Recent efforts~\cite{watch_it_move,yao2022lassie} aim to build articulated 3D object models without templates or category-specific priors. However, they either require synchronized multi-view videos which are hard to acquire, or a manually created skeleton as initialization.

In this work, we propose a novel approach to build an animatable 3D model for any articulated object using only monocular videos. We do not rely on prior knowledge of the object shape and structure or manual annotations as initialization. Thus, our pipeline significantly reduces the amount of work needed to create animatable 3D models and eliminates the need for synchronized multi-view observations or specialized sensors. Our model can be directly used for pose manipulations and animations in 3D, which we refer to in this paper as \emph{animatable model}.

Specifically, our approach uses Neural Radiance Fields~\cite{mildenhall2020nerf} and Signed Distance Functions to represent the appearance and shape of the object. The body pose of the object is represented by a \emph{kinematic chain}. Our kinematic chain is initialized based on the initial estimate of the object's shape, and further optimized jointly with the object's shape, appearance, and deformation parameters. We adopt skinning weights mechanisms from~\cite{saito2021scanimate,chen2021snarf,banmo} and propose a novel way to use our optimized kinematic chain to drive pose changes and deformations. Due to the under-constrained nature of this problem, we leverage the foreground masks and optical flow predicted by off-the-shelf models as robust visual cues to learn the shape and underlying structure of the object. Inspired by~\cite{yao2022lassie}, we utilize the 2D image features from DINO-ViT~\cite{DINO} that are trained in a self-supervised manner, to establish long-range correspondences and consistencies at the object parts level~\cite{amir2021deep,tumanyan2022splicing} between different frames. This is achieved by matching canonical features at the object surface to the pre-trained DINO-ViT image features. Examples of rendered objects in canonical representations and novel poses, along with the corresponding 3D meshes and kinematic chains are shown in~\cref{fig:teaser}. Our contributions are summarized as follows:
\begin{itemize}
    \item Our work is the first to build an animatable 3D model for objects of arbitrary categories from monocular videos without any template or prior knowledge of the object's shape and underlying structure.
    \item We propose a novel optimization technique to iteratively refine the kinematic chain and its associated deformation parameters. Users can directly manipulate the optimized kinematic chain to animate the object.
    \item We achieve similar surface reconstruction results to state-of-the-art 3D surface reconstruction methods on various articulated and deformable object categories.
\end{itemize}
\section{Related Work}
\label{sec:related_works}

\begin{figure*}[t!]
    \centering
    \includegraphics[width=\linewidth, trim={0 0 0 0},clip]{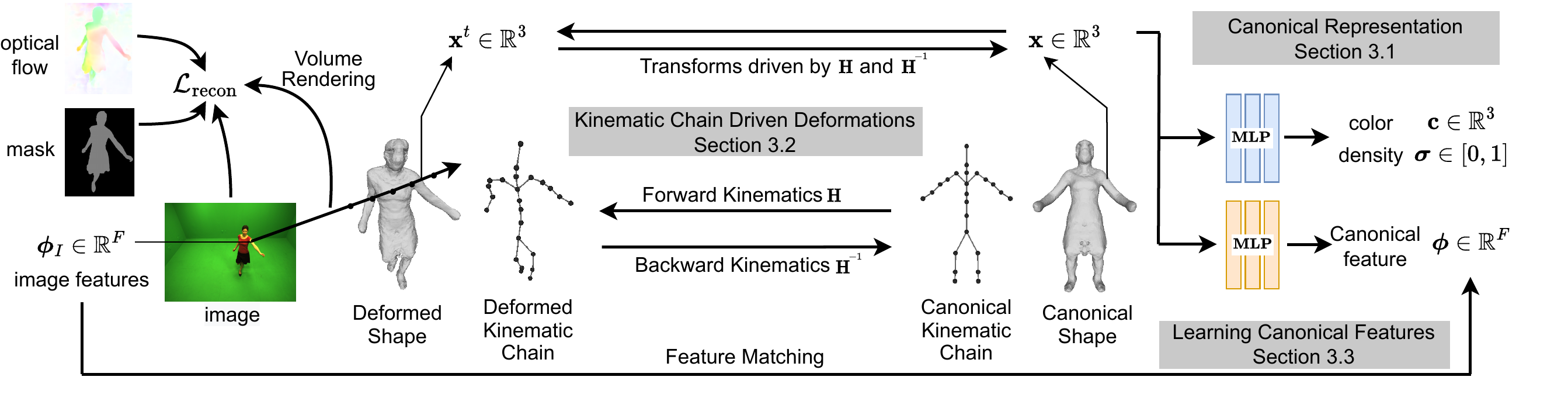}
    \caption{\textbf{Method overview}. Our method optimizes the canonical representation, including the object's shape, appearance, and kinematic chain~(\cref{sec:canonical_representation}), to enable direct pose manipulations. It uses the kinematic chain to transform 3D points between the canonical and deformed space~(\cref{sec:kinematic_chain_driven_deform}), and render 2D observation predictions of colors, foreground masks and optical flows to match the actual 2D observations. In addition, canonical feature embeddings~(\cref{sec:canonical_feature}) are learned by matching the corresponding 2D image features to establish the object parts level correspondences across different frames within a collection of monocular videos.
    }
    \label{fig:overview}
\vspace{-10pt}
\end{figure*}

\noindent\textbf{3D deformable shape reconstruction from images}. Inspired by NeRF~\cite{mildenhall2020nerf}, recent works~\cite{li2021neural,pumarola2021d,tretschk2021non,wang2021neural,kong2019deep,park2021nerfies} are able to obtain robust 3D reconstructions on dynamic scenes or deformable objects from a collection of multi-view images. Some works\cite{kanazawa2018learning,kolotouros2019learning,ye2021shelf,li2020online} use additional supervisions from human annotated data or shape templates to recover 3D shapes. These methods suffer from large performance degradation when the input has large deformations and self-occlusions. Recent works~\cite{kumar2020non,sidhu2020neural,zeng2021pr} explore structure from motion approaches to reconstruct non-rigid 3D scenes using video sequences. Similar to us, LASR\cite{yang2021lasr}, ViSER\cite{yang2021viser}, and BANMo~\cite{banmo} utilize monocular videos for reconstructing deformable and articulated 3D shapes and achieve great surface reconstruction results. However, these approaches do not suffice the needs of direct pose manipulations of the reconstructed 3D objects as they reconstruct by memorizing the poses and movements in training data.

\noindent\textbf{Category-specific animatable models}. A group of works~\cite{chan2019everybody, liu2019liquid, ma2017pose, yoon2021pose, zhu2019progressive} tackle the problem of animating humans in 2D using annotated data with human poses. To enable animations in 3D space, significant efforts from the community have been put into creating 3D human and animal shape models~\cite{loper2015smpl,pavlakos2019expressive,xiang2019monocular,xu2020ghum}. Many  works~\cite{vo2020spatiotemporal,peng2021neural,deng2020nasa,noguchi2021neural,weng2022humannerf,peng2021animatable,liu2021neural,kundu2020appearance, wang2022arah,wu2022casa,wu2022magicpony,huang2020arch,he2021arch++,xu2019denserac} utilize these template shapes or pose priors from shape models to recover 3D shapes and perform animations. Recent methods~\cite{su2021nerf,su2022danbo,li2022tava} do not use shape templates, but rely on pre-defined skeletons or poses predicted by models~\cite{kocabas2020vibe,kolotouros2019learning} trained on human annotated data, and these methods require synchronized multi-view inputs. These pre-defined skeletons or poses are used as priors for learning the shape and articulation for animations. However, the annotations are usually expensive to get and category-specific, which cannot be applied to other object categories in the real world and do not work for out-of-distribution articulation modes or poses. 

\noindent\textbf{Category-agnostic animatable models}. Several methods have recently emerged for building 3D animatable model for any articulated object. Watch it Move~\cite{watch_it_move} builds a 3D articulated structure by identifying physically meaningful joints in an unsupervised manner from calibrated multi-view videos and corresponding foreground masks. Recent work~\cite{tang2022neural} learns deformation behaviours from a given 3D mesh of the object to allow users to define desired deformations at regions of interest. However, synchronized multi-videos and 3D meshes are not easily accessible. Similar to our goal, LASSIE~\cite{yao2022lassie} recovers a 3D shape and skeleton from image ensembles without pre-defined shape priors, but it requires a manually annotated 3D skeleton as input. In contrast, we aim to simplify the process of building animatable 3D models for any articulated object using only collections of monocular videos that can be easily collected.

\section{Method}
\label{sec:method}

Given a collection of monocular videos, our goal is to build an animatable 3D model for the articulated object in the scene. Our method optimizes the canonical representation~(\cref{sec:canonical_representation}) of the object, where its shape and appearance are modeled implicitly using Multi-layer Perceptrons (MLPs). Its canonical body pose is represented by a kinematic chain. Our approach consists of two optimization stages: the first stage~(\cref{sec:init_opt}) optimizes the canonical shape and deformation parameters in each frame without a kinematic chain to get an initial estimate of the 3D shape. We then apply RigNet~\cite{xu2020rignet}, an off-the-shelf category-agnostic skeleton estimator, on the initial estimate of the 3D mesh to obtain an initial kinematic chain. In the second stage~(\cref{sec:kinematic_chain_opt}), the kinematic chain adapts to object shape to enable kinematic chain driven deformations~(\cref{sec:kinematic_chain_driven_deform}). We supervise the learning by matching the rendered images, foreground masks, and optical flows to the actual 2D observations in both stages. In addition, we learn canonical feature embeddings~(\cref{sec:canonical_feature}) to obtain long-range correspondences between video frames, supervised by pre-trained 2D image features from~\cite{DINO}.

\subsection{Canonical Representation}
\label{sec:canonical_representation}
\noindent\textbf{Canonical shape}.
Similar to~\cite{banmo,mildenhall2020nerf}, we represent the shape and appearance of an object implicitly in the canonical space. 
A 3D point $\mathbf{x} \in \mathbb{R}^3$ has two properties: its color $\mathbf{c} \in \mathbb{R}^3$ and density $\boldsymbol{\sigma} \in [0,1]$. The color $\mathbf{c}$ is given by a Multilayer Perceptron (MLP) network, and the density $\boldsymbol{\sigma}$ is given by the cumulative of a Laplacian distribution with zero mean and learnable scale on the learned Signed Distance Function (SDF)~\cite{yariv2021volume,wang2021neus} in the canonical space. The value of SDF at any 3D point is given by a separate MLP network, and the canonical mesh can be extracted by finding the zero-level set of the SDF~\cite{yariv2021volume} with the marching cubes algorithm. Please see our supplementary material for additional details about the MLPs.

\noindent\textbf{Canonical kinematic chain}.
We use a kinematic chain to represent the body pose of any articulated object. The canonical kinematic chain is defined as a set of connected 3D joints $\mathbf{P}=\left\{\mathbf{p}_{i} \mid i=1, \ldots, {n}_{p}\right\}$, where $\mathbf{p}_{i} \in \mathbb{R}^{3}$ denotes joint positions in the canonical space, with a set of ${n}_{j}-1$ links $\mathbf{L}=\left\{\mathbf{\boldsymbol {\ell}}_{jk}=(\mathbf{p}_{j},\mathbf{p}_{k})\right\}^{({n}_{j}-1)}$. The joints are connected in a hierarchical manner with a pre-defined root joint to form a tree structure. As a result, there is no cycle, and a unique path exists between every two joints.

\subsection{Kinematic Chain Driven Deformations}
\label{sec:kinematic_chain_driven_deform}
\subsubsection{Neural Blend Skinning Deformations}

Inspired by~\cite{yang2021lasr,yang2021viser,banmo}, we define a set of learnable \emph{deformation anchors} $\mathbf{A}=\left\{\mathbf{a}_{i} \mid i=1, \ldots, {n}_{a}\right\}$, where $\mathbf{a}_{i} \in \mathbb{R}^{3}$ denotes the anchor positions in the canonical space. We compute surface deformations with respect to anchors via linear blend skinning with learned weights. At time $t$, let ${\bf C}^t \in SE(3)$ be the object's root pose with respect to the canonical root pose, ${\bf T}^t_{i} \in SE(3)$ be the transformation of anchor $\mathbf{a}_i$ from canonical space to the deformed space, and $\mathbf{w}^t_{{a}_i}$ be the skinning weight of $\mathbf{x}$ with respect to anchor $\mathbf{a}_i$. Given any 3D point ${\mathbf{x}} \in \mathbb{R}^{3}$ in the canonical space, its corresponding point in the deformed space $\mathbf{x}^t$ at time $t$ is given by the weighted average of anchor's transformation ${\bf T}^t_{i}$ by: 
\begin{equation}
\label{eq:forward_skinning}
{\mathbf{x}}^{t} = {\mathbf{C}}^{t} \sum_{i=1}^{{n}_{a}} ({\mathbf{w}}_{{a}_{i}} {\bf T}^t_{i}) \mathbf{x}
\end{equation}
where we define $\mathbf{W} = \left[ \mathbf{w}_{{a}_1}, \dots, \mathbf{w}_{{a}_{{n_a}}}  \right] \in \mathbb{R}^{n_a}$ to be the forward skinning weights for canonical space point $\mathbf{x}$ at time $t$, and the weights are determined by the Euclidean distances between point $\mathbf{x}$ and all anchors $\mathbf{A}$ in the canonical space as:
\begin{equation}
    \label{eq:forward_weights}
    {\bf W} = \sigma \bigg (-\left \| \mathbf{x} - \mathbf{A} \right \|^2_2 \bigg )
\end{equation}
where $\sigma$ is softmax to normalize the skinning weights and its temperature $\tau$ is a learnable parameter.

\subsubsection{Kinematic Chain Driven Deformations}
\begin{figure}[t!]
    \centering
    \includegraphics[width=\linewidth, trim={0cm 0.1cm 0cm 0.2cm},clip]{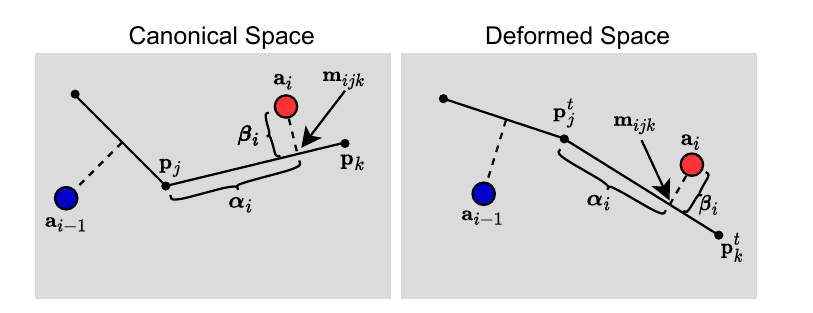}
    \caption{Simplified 2D example on how anchors (red and blue dots) move along with associated kinematic chain links. Associations can be visualized at dashed lines.}
    \label{fig:associoation}
    \vspace{-10pt}
\end{figure}

As the kinematic chain represents the body pose of the object, the deformation anchors' transformations must be driven by the kinematic chain to obtain properly deformed surfaces corresponding to the actual body pose. To achieve this, we associate each anchor $\mathbf{a}_{i}$ to its closest kinematic chain link $\mathbf{\boldsymbol {\ell}}_{jk}$ based on Euclidean distances in the canonical space. Example associations are shown as dashed lines in~\cref{fig:associoation}. We let the intersection between the dashed line of anchor $\mathbf{a}_i$ and the associated link $\mathbf{\boldsymbol {\ell}}_{jk}$ be $\mathbf{m}_{ijk}$, where the parent joint and child joint of link $\mathbf{\boldsymbol {\ell}}_{jk}$ are $\mathbf{p}_j$ and $\mathbf{p}_k$. Then this association between anchor $\mathbf{a}_i$ and link $\mathbf{\boldsymbol {\ell}}_{jk}$ can be represented by the following terms:

\begin{equation}
    {\boldsymbol{ \alpha}}_i = \left \| \mathbf{m}_{ijk} - \mathbf{p}_j \right \|_2, \space {\boldsymbol{ \beta}}_i = \left \| \mathbf{a}_i - \mathbf{m}_{ijk} \right \|_2, \notag
\end{equation}
\begin{equation}
    \label{eq:asso_all}
    {\mathbf{G}_i} = \mathbf{Rot} (\mathbf{m}_{ijk} - \mathbf{p}_j, \mathbf{a}_i - \mathbf{m}_{ijk}),
\end{equation}
where $\boldsymbol{\alpha}_i$ is the Euclidean distance between the parent joint $\mathbf{p}_j$ and the intersection $\mathbf{m}_{ijk}$, and $\boldsymbol{\beta}_i$ is the Euclidean distance between the intersection $\mathbf{m}_{ijk}$ and associated anclhor $\mathbf{a}_i$. The rotation matrix $\mathbf{G}_i \in SO(3)$ aligns vectors $\mathbf{m}_{ijk} - \mathbf{p}_j$ and $\mathbf{a}_i - \mathbf{m}_{ijk}$ about the intersection $\mathbf{m}_{ijk}$. We refer to these three terms as the \emph{association parameters} for anchor $\mathbf{a}_i$, and they are kept constant in \emph{any} kinematic chain configuration to ensure stable surface deformations.

With defined association parameters, we can express an anchor's position using its associated link and association parameters. As anchor $\mathbf{a}_i$ is associated to link ${\ell}_{jk}$ in the canonical space, its universal position $\mathbf{\tilde a}_i$ in \emph{any} kinematic chain configuration can be computed as:
\begin{equation}
    \label{eq:deformed_anchors}
    {\mathbf{\tilde{a}}_i} = \mathbf{\tilde p}_j + {\underbrace {\boldsymbol{\alpha}_i \frac{\mathbf{\tilde p}_{k} - \mathbf{\tilde p}_{j}}{\left \| \mathbf{p}_{k} - \mathbf{p}_j \right \|_2}}_\text{parent joint to intersection}} + {\underbrace {\boldsymbol{\beta}_i \mathbf{G}_i \frac{\mathbf{\tilde p}_{k} - \mathbf{\tilde p}_{j}}{\left \| \mathbf{p}_{k} - \mathbf{p}_j \right \|_2}}_\text{intersection to anchor}},
\end{equation}
where $\mathbf{\tilde p}_j$ and $\mathbf{\tilde p}_k$ are given by:
\begin{equation}
  \mathbf{\tilde p}_{j,k}=\begin{cases}
    \mathbf{p}_{j,k} & \text{in canonical space}\\
    \mathbf{H}_{j,k}^t \mathbf{p}_{j,k} & \text{in deformed space at time }t
  \end{cases}
\end{equation}
Note that $\mathbf{\tilde{a}}_i = \mathbf{a}_i$ in the canonical space.  $\mathbf{H}_{j}^t \in SE(3)$ and $\mathbf{H}_{k}^t \in SE(3)$ are the rigid transformations of joint $\mathbf{p}_j$ and joint $\mathbf{p}_k$ from the canonical space to the deformed space without breaking the kinematic chain, which we refer as \emph{forward kinematics} on the kinematic chain.

Then the rigid transformation $\mathbf{T}_i^t$ of anchor $\mathbf{a}_i$ at time $t$ can be directly inferred based on its new position in the deformed space given by~\cref{eq:deformed_anchors}:
\begin{equation}
    \label{eq:revised_anchors_transforms}
    {\mathbf{T}_i^t} = \left[ \mathbb{I} \mid \mathbf{\tilde{a}}_i - \mathbf{a}_i  \right],
\end{equation}
where $\mathbb{I} \in \mathbb{R}^{{3} \times {3}}$ is the identity matrix.

Given optimized kinematic chain and anchors, we can perform explicit re-posing of the object by directly applying user defined forward kinematics $\mathbf{H}$ to the kinematic chain.

\subsection{Learning Semantically Consistent Features}
\label{sec:canonical_feature}
Large deformations and self-occlusions often cause 3D reconstruction methods to fail or perform poorly. This is usually due to the lack of strong correspondence cues or the lack of long-range correspondences between different frames in long videos~\cite{sand2008particle, sundaram2010dense}. Recent works~\cite{amir2021deep,tumanyan2022splicing} show that the pre-trained image features from DINO-ViT~\cite{DINO, ViT} can provide robust correspondences at the object parts level between different 2D images even under large appearance variations and viewing angle changes. 

To better optimize the canonical shape and deformations, we follow~\cite{yang2021viser,banmo} to learn surface feature embeddings $\phi(\mathbf{x}) \in \mathbb{R}^{F}$ for points on the canonical shape's surface. The feature is given by a separate MLP network that takes in any 3D point in the canonical space and it is trained in a self-supervised manner such that the surface feature embeddings match the pre-trained 2D image features at its corresponding pixel given by the camera projection model. Compared to BANMo~\cite{banmo} where pre-trained DensePose CSE features~\cite{neverova2020continuous} specifically designed for humans and animals are used for learning surface embedding, DINO-ViT~\cite{DINO} features have shown to be meaningful and rich in downstream tasks for a wide variety of objects~\cite{DINO,amir2021deep}. This is because DINO-ViT is pre-trained in a self-supervised manner on large-scale datasets with diverse object categories. 

\subsection{Two-Stage Optimizations}

\subsubsection{Initial Optimization}
\label{sec:init_opt}
The initial optimization stage aims to build a reasonable object shape with unconstrained anchors to capture deformations in each frame of the training videos. We initialize the MLP network that predicts the canonical shape of the object to approximate the object as a 3D ellipsoid, and initialize all the anchors $\mathbf{A}$ at the global origin in the canonical space. During the initial optimization stage, the anchors are updated without any constraints in terms of their positions. Inspired by~\cite{banmo}, the \emph{unconstrained transformations} of the anchors at time $t$ are given by a separate MLP, $\mathbf{\mathcal{F}_{A}}$ as:
\begin{equation}
\label{eq:anchor_transforms}
{\bf \hat T}^t = \mathbf{\mathcal{F}_A}(\boldsymbol\psi_a^t),
\end{equation}
where $\boldsymbol\psi_a^t \in \mathbb{R}^{128}$ is a learnable latent code for time $t$. 

As we define the transformation of any 3D point $\mathbf{x}$ from the canonical space to the deformed space as weighted rigid transformation of anchors in~\cref{eq:forward_skinning}, we can directly infer the backward transformation for any 3D point $\mathbf{x}^t$ in the deformed space by taking the inverse of the rigid transformations as:
\begin{equation}
    \mathbf{x} = \bigl ( \sum_{i=1}^{{n}_{a}} \bigl ( {\mathbf{w}}_{{a}_{i}}^t ({\bf \hat T}^t_{i})^{-1} \bigr) \bigr ) {(\mathbf{C}}^{t})^{-1} \mathbf{x}^t \notag
\end{equation}
\begin{equation}
    \label{eq:backward_weights}
    {\bf W^t} = \sigma \biggl ( -\left \| \mathbf{x}^t - \mathbf{A}^t \right \|^2_2 \bigg ),
\end{equation}
where the backward skinning weights $\mathbf{W^t} = \left[ \mathbf{w}_{{a}_1}^t, \dots, \mathbf{w}_{{a}_{{n_a}}}^t  \right] \in \mathbb{R}^{n_a}$ are computed in the same way as in forward skinning in~\cref{eq:forward_weights} but based on deformed anchors and the 3D point of interest in the deformed space instead. This allows us to directly transform any 3D point back and forth between the canonical space and each frame's deformed space.

Similar to~\cite{yang2021viser,banmo}, the canonical shape and appearance are learned by minimizing the reconstruction losses on 2D observations: RGB images, foreground masks, and optical flow. The reconstruction losses are formulated in the same way as in~\cite{mildenhall2020nerf,banmo,yang2021viser,yang2021lasr,yariv2020multiview}, where we minimize the differences between the rendered and the actual observations:
\begin{equation}
    \label{eq:reconstruction_losses}
    \mathcal{L}_{\textrm{recon}} = \sum_{\mathbf{x}_I} \left \| \mathbf{o}_{r} - \mathbf{o}_{gt} \right \|^2,
\end{equation}
where $\mathbf{o}_{r}$ and $\mathbf{o}_{gt}$ are the pairs of rendered and ground-truth 2D observations (2D images, foreground masks, optical flow) at pixels of interest $\mathbf{x}_I^t \in \mathbb{R}^2$ at time $t$. To perform volumetric rendering at time $t$, we first sample a camera ray starting at the pixel of interest on the 2D image at time $t$, and use~\cref{eq:backward_weights} to transform the sampled points along the camera ray back to the canonical space. We then follow the same rendering process in NeRF~\cite{mildenhall2020nerf} on the camera ray in the canonical space to get the colour $\mathbf{c}^t$ at the pixel of interest. To render optical flow, we follow~\cite{banmo} to transform canonical space points to consecutive frames' deformed spaces, and use the camera projection model to find their corresponding positions on the 2D image. We compute the difference in the pixel locations as the optical flow.

The anchors and their transformations are learned from scratch with additional losses to enforce cycle consistency. The consistency losses enforce any 3D point in the deformed space after a backward transformation and a forward transformation to end up at the same position. To supervise canonical feature embeddings, we apply soft argmax descriptor matching~\cite{kendall2017end,luvizon2019human, yang2021viser, banmo} on the 2D pixel of interest to determine the most probable corresponding surface point in the canonical space. The matching is based on cosine similarities between the canonical features and the pre-trained 2D image features. A 2D consistency loss~\cite{banmo} is used to minimize the difference between matched position and the location given by the camera projection model. The details of loss functions are in the supplementary material.

\subsubsection{Kinematic chain aware optimization}
\label{sec:kinematic_chain_opt}
After the initial optimization stage, we obtain an estimate of the canonical shape and a set of unconstrained anchors. However, the unconstrained anchors introduce unrealistic artifacts in deformations, such as undesired shrinks and stretches on the canonical shape. And the anchors are only optimized to ``mimic'' object shape in each frame of the videos. A kinematic chain is needed to remove these undesired effects and enable explicit re-posing of the object.

We initialize our canonical kinematic chain with pre-trained RigNet~\cite{xu2020rignet} that takes in our canonical mesh from the initial optimization stage and outputs a set of joints and connections. The pre-trained RigNet is able to provide a reasonable estimate of the kinematic chain. However, it requires further optimization such that our kinematic chain can better adapt to the object's movements in the training videos, and therefore align itself with the object's actual underlying structure in each frame.

As any point in 3D can be transformed back and forth between the canonical space and the deformed space using the transformations defined by anchors, we utilize the learned unconstrained anchor transformations $\mathbf{\hat T}^t$ from the initial optimization stage to transform the canonical kinematic chain joints $\mathbf{P}$ following~\cref{eq:forward_skinning}:
\begin{equation}
\label{eq:forward_kinematic_chain}
{\mathbf{\hat P}}^{t} = {\mathbf{C}}^{t} \sum_{i=1}^{{n}_{a}} ({\mathbf{w}}_{{a}_{i}} {\bf \hat T}^t_{i}) \mathbf{P},
\end{equation}
where $\mathbf{\hat P}^t=\left\{\mathbf{\hat p}_{i} \mid i=1, \ldots, {n}_{p}\right\}$ is the set of \emph{unconstrained kinematic chain joints} in the deformed space at time $t$. The transformations of anchors are learned without any regularization on how they move or any knowledge on the existence of the kinematic chain. Therefore the transformed joints $\mathbf{\hat P}^t$ computed using unconstrained anchors will break the kinematic chain. We can recover the properly deformed kinematic chain as a set of \emph{revised joints} $\mathbf{\tilde P}^t=\left\{\mathbf{\tilde p}_{i} \mid i=1, \ldots, {n}_{p}\right\}$ in the deformed space. It is done by updating each unconstrained joint's position in a hierarchical order to enforce the length of each kinematic chain link to be constant. Note that the order of joint connections is preserved at all times after initialization to maintain the hierarchical structure of the kinematic chain and stability during optimization. The detailed steps to recover the kinematic chain are explained in the supplementary material.

As we have the fixed associations between anchors and the initial kinematic chain built in the canonical space as $\boldsymbol{\alpha}_i$, $\boldsymbol{\beta}_i$, and $\mathbf{G}_i$ by~\cref{eq:asso_all}. We then use~\cref{eq:deformed_anchors} to recover each anchor's revised position $\mathbf{\tilde a}_i$ and thus apply~\cref{eq:revised_anchors_transforms} to infer a revised anchor transformation $\mathbf{\tilde T}_i^t$ that satisfies the association constraints between the anchors and the kinematic chain, while not breaking the kinematic chain at the same time. During the optimization, we introduce a novel loss that minimizes the differences between the unconstrained anchors and the revised anchors:
\begin{equation}
    \label{eq:lookup_table_loss}
    \mathcal{L}_{\textrm{anchors}} = \sum_{i=1}^{n_a} \left \| \mathbf{\tilde a}_i - \mathbf{\hat T}_i^t \mathbf{a}_i \right \|^2_2.
\end{equation}
This loss aims to make the optimized deformation anchors and the MLP network $\mathbf{\mathcal{F}_{A}}$ kinematic chain aware.

Additionally, we optimize the kinematic chain by introducing a learnable \emph{additive residual} term for each kinematic chain link. The length of each link in the kinematic chain is updated using the learnable residual during optimization such that the kinematic chain can better adapt to the object's shape and the learned deformation anchors. The details on how the kinematic chain is updated without breaking itself are included in the supplementary material.

In the kinematic chain aware optimization stage, all the MLP networks are also optimized jointly to finetune the object's shape and appearance with the loss functions in the initial optimization stage. At the end of this stage, a fully controllable articulated model for the object is built, and animations of the object can be easily done by directly manipulating the final kinematic chain. 
\section{Experiments}
\label{sec:experiments}

\subsection{Datasets}
\noindent\textbf{Synthetic Datasets}.
We validate our approach on two synthetic datasets. The Eagle~\cite{banmo} dataset consists of 5 single-view videos (900 frames in total) of an animated eagle model from Turbosquid. We also created a new dataset (iiwa dataset) that contains 4 single-view videos (1200 frames in total) of an animated iiwa robot arm model from BlenderKit. Please find the details and examples of our iiwa dataset in the supplementary material.

\noindent\textbf{AMA Dataset}. We also evaluate our approach on the Articulated Mesh Animation (AMA) Dataset~\cite{vlasic2008articulated}. It contains 10 sets of multi-view videos of 3 different clothed humans using synchronized cameras. We use 2 challenging sets of videos and treat the multi-view videos as a collection of monocular videos by discarding the time synchronization information, which yields 2600 monocular frames in total. 

\noindent\textbf{In-the-wild Datasets}. In addition, we validate our approach on in-the-wild monocular captures of various deformable and articulated objects
datasets. In the main paper, we focus on experimental results on the synthetic and AMA datasets. Please see the supplementary material for implementation details and results on in-the-wild datasets. 

\begin{figure*}[t!]
  \centering
  \includegraphics[width=1\textwidth]{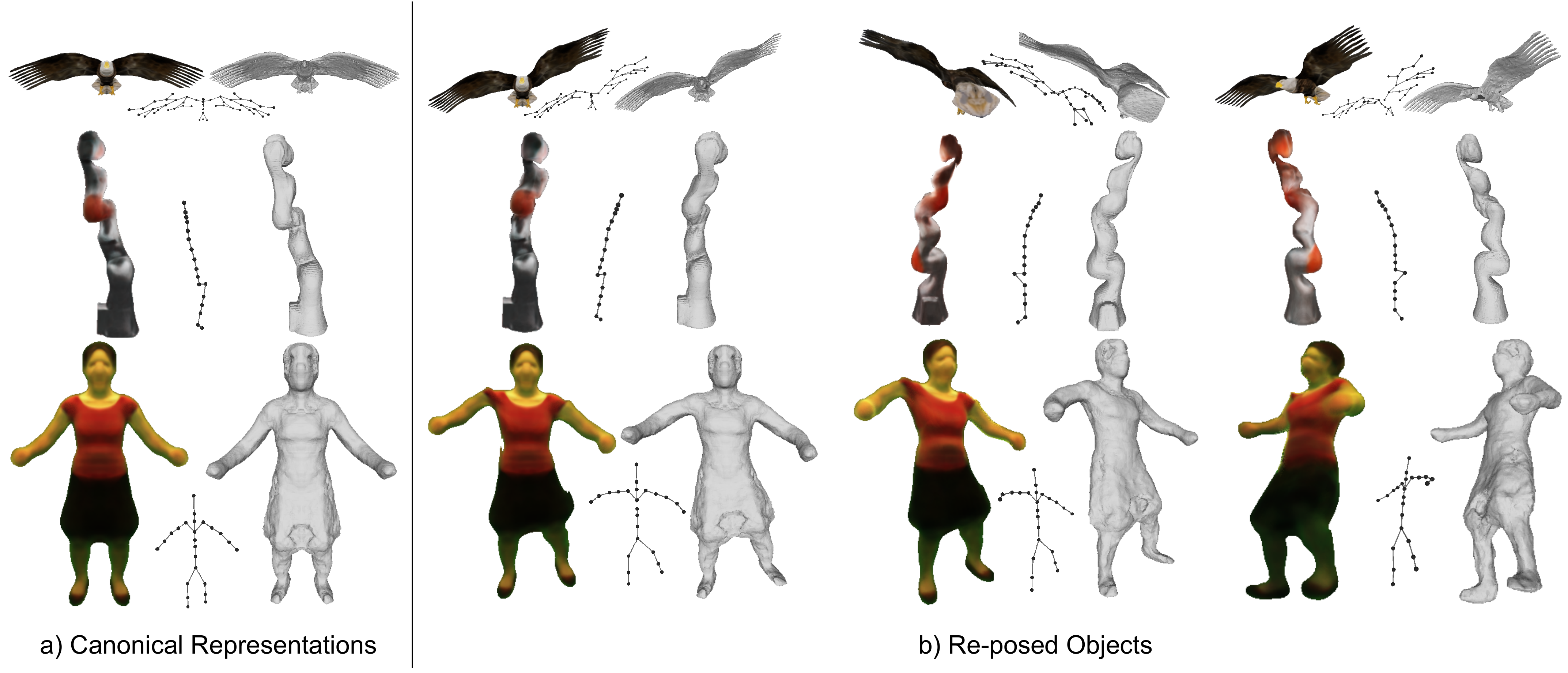}

  \caption{\textbf{Pose manipulation examples}. In a) we show the learned canonical representations of the object for each dataset. We perform pose manipulations using optimized kinematic chain and show the transformed kinematic chain, re-posed mesh as well as the rendered object in three different views in b).}
  \label{reposing_results}
  \vspace{-10pt}
\end{figure*}

\subsection{Implementation Details} For synthetic and AMA datasets, we use ground-truth foreground masks during the optimization. For root pose initializations, we use ground-truth root poses for synthetic datasets, while a pre-trained PoseNet~\cite{banmo} is used to initialize the root poses for AMA Dataset. The root poses are also updated during optimization with learned residual terms for AMA Dataset. The optical flows for all datasets are computed by VCN~\cite{yang2019vcn} at frame intervals of $\left \{ 1, 2, 4, 8, 16, 32 \right \} $ in both forward and backward time directions. We use the \emph{keys} of the last transformer layer from a pre-trained ViT-S/8 model~\cite{DINO} as supervisions for the canonical features. We perform PCA on the 2D image features to reduce the dimension from 384 to 16, to match our canonical feature's dimension. To generate mesh from our implicit representations of the object shape, we run marching cubes on $256 \times 256 \times 256$ grids to extract the zero-level set of the learned SDF. We use 10, 25, and 36 anchors for iiwa, Eagle, and AMA Dataset, respectively. Our code and iiwa dataset will be made publicly available. 

\subsection{Pose Manipulation Results}

The advantage of having a kinematic chain is that users can manipulate the 3D object to poses that had never occurred in training videos. Although some surface reconstruction methods~\cite{banmo,yang2021viser} allow users to move their learned ``floating'' control points around for custom animations, these control points are not intuitive to manipulate because it's unclear how each ``floating'' control point contributes to the overall deformation. Moreover, these ``floating'' control points are optimized for only memorizing the poses and movements in training videos. In our approach, the pose manipulations are done by directly applying rotations at kinematic chain joints to rotate and twist corresponding kinematic chain links, which is much more straightforward and intuitive for animation purposes. We do not impose any constraint on the kinematic chain configuration as long as the order of joint connections and the length of each link are preserved. Users are expected to apply valid transformations to the kinematic chain to obtain feasible shapes without mesh degeneration. We show the learned canonical shapes, optimized kinematic chains, and the re-posed objects in novel poses that do not occur in training videos in~\cref{reposing_results}. Please see our supplementary material for additional re-posing results.

\begin{figure}[t!]
    \centering
    \includegraphics[width=\linewidth, trim={0cm 0.1cm 0cm 0.2cm},clip]{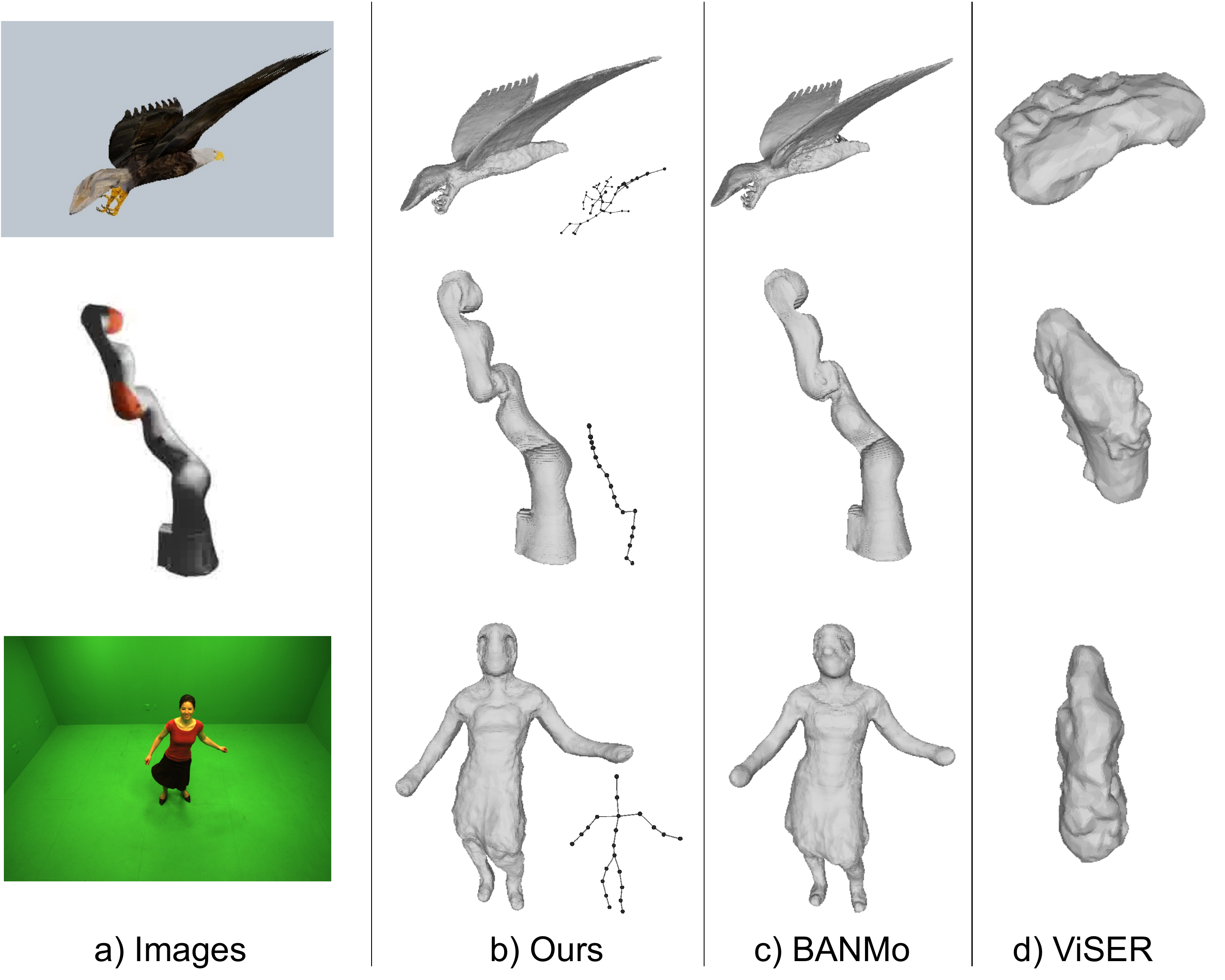}
    \caption{\textbf{Qualitative comparision of our method with BANMo~\cite{banmo} and ViSER~\cite{yang2021viser}.}} 
    \label{fig:reconstruction_figs}
    \vspace{-10pt}
\end{figure}

\subsection{Reconstruction Results}
There is no prior work on building category-agnostic and animatable 3D models from monocular videos, and reconstruction quality is an important factor for realistic novel view synthesis and animations. Therefore we evaluate the performance of our method on 3D surface reconstruction using ground-truth 3D meshes provided by each dataset. 

\noindent\textbf{Metrics}.
We align the reconstructed 3D mesh to the ground-truth mesh using Iterative Closest Point (ICP) if they are in different scales and orientations before evaluation. Following~\cite{groueix2018papier,fan2017point,sun2018pix3d,banmo}, we report 3D Chamfer Distances between reconstructed mesh vertices and the ground-truth mesh vertices measured in centimeters. In addition, we follow~\cite{tatarchenko2019single} to report the F-scores at a distance threshold of 2$\%$. All the reported numbers are averaged across all frames for each dataset. In general, these metrics measure the following two aspects of the reconstruction quality for each frame in the training videos: the overall quality of the reconstructed 3D mesh surface, and whether the model is able to predict an accurate shape corresponding to the correct object body pose.

\noindent\textbf{Baselines}.
We compare our approach with two baselines in 3D surface reconstruction for deformable objects: BANMo~\cite{banmo} and ViSER~\cite{yang2021viser}. We provide the same root pose initializations to all the methods for fair comparison on each dataset. Note that BANMo and ViSER \emph{cannot} perform direct pose manipulations, and BANMo utilizes DensePose CSE feature embeddings~\cite{neverova2020continuous} designed for humans and quadruped animals, which makes it a category-specific method. We show qualitative comparision of our method with BANMo~\cite{banmo} and ViSER~\cite{yang2021viser} in~\cref{fig:reconstruction_figs}. Our approach can achieve similar 3D surface reconstruction quality using a category-agnostic approach while enabling direct pose manipulations. We also show the quantitative comparison with the same baselines: BANMo~\cite{banmo} and ViSER~\cite{yang2021viser} on each dataset in~\cref{tab:reconstruction_comparision}.

\begin{table}
\centering
\caption{3D surface reconstruction results evaluated in 3D Chamfer Distances ($\downarrow$) and F-scores ($\uparrow$) at a distance threshold of 2$\%$ averaged across all frames.}
\resizebox{\columnwidth}{!}{
\begin{tabular}{c|cc|cc|cc}
\label{tab:reconstruction_comparision}
\multirow{2}{*}{Method} & \multicolumn{2}{c|}{Eagle}     & \multicolumn{2}{c|}{iiwa}      & \multicolumn{2}{c}{AMA-swing}   \\ 
\cline{2-7}
                        & CD            & F@2\%          & CD            & F@2\%          & CD            & F@2\%           \\ 
\hline
ViSER                   & 32.6          & 9.4            & 20.6          & 18.1           & 17.9          & 39.2            \\
BANMo                   & 4.44          & 82.72          & 5.55          & 54.58          & \textbf{9.28} & \textbf{56.24}  \\
Ours                    & \textbf{4.21} & \textbf{83.38} & \textbf{5.52} & \textbf{56.53} & 9.69          &  53.29 
\end{tabular}
}
\vspace{-20pt}
\end{table}

\subsection{Ablations}
We run ablation studies on important components of our method to evaluate their effects. Please refer to our supplementary materials for more comprehensive results.

\begin{figure}[t!]
    \centering
    \includegraphics[width=\linewidth, trim={0cm 0.1cm 0cm 0.2cm},clip]{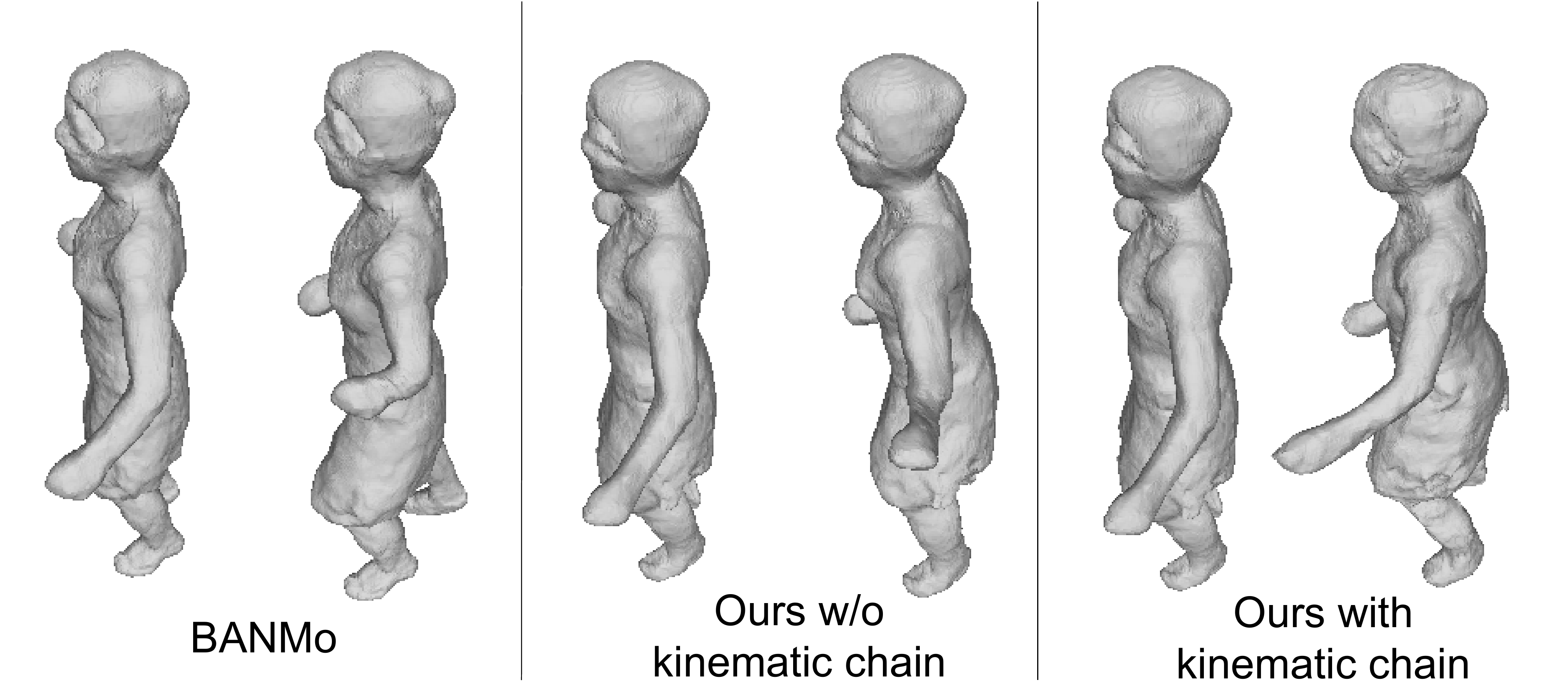}
    \caption{\textbf{Ablation on kinematic chain}. We show canonical shapes~(left) and the predicted shapes~(right) in the same time instance for each method. The length of the left arm is preserved with the help of kinematic chain.}
    \label{fig:kinematic_chain_ablation}
    \vspace{-10pt}
\end{figure}

\noindent\textbf{Kinematic chain}. A kinematic chain enables explicit pose manipulations and effectively reduces the unrealistic surface deformations that often occur in 3D surface reconstruction methods. We show the improvements of having a kinematic chain in~\cref{fig:kinematic_chain_ablation}. The kinematic chain correctly re-poses the person's arm without volume loss, while BANMo~\cite{banmo} and our method with unconstrained anchors fail to model the correct deformations. We also compare the reconstruction results of directly using kinematic chain initialization from~\cite{xu2020rignet} and using optimized kinematic chain in~\cref{tab:kinematic_chain_ablation}. We are able to gain considerable improvements with the kinematic chain aware optimization stage~(\cref{sec:kinematic_chain_opt}) because the optimized kinematic chain aligns better with the underlying object body pose in each frame.

\begin{table}
\centering
\caption{Ablation on kinematic chain aware optimization evaluated in 3D Chamfer Distances ($\downarrow$) and F-scores ($\uparrow$) at a distance threshold of 2$\%$ averaged across all frames.}
\resizebox{\columnwidth}{!}{
\begin{tabular}{c|cc|cc|cc}
\label{tab:kinematic_chain_ablation}
\multirow{2}{*}{Method} & \multicolumn{2}{c|}{Eagle}     & \multicolumn{2}{c|}{iiwa}      & \multicolumn{2}{c}{AMA-swing}   \\ 
\cline{2-7}
                        & CD            & F@2\%          & CD            & F@2\%          & CD            & F@2\%           \\ 
\hline
w/o opt.                   & 7.44          & 57.87          & 5.69          & 52.44          & 11.13 & 48.88  \\

with opt.                    & 4.21 & 83.38 & 5.52 & 56.53 & 9.69          &  53.29  \\
\hline 
Improv. & -3.23 & +25.51 & -0.17 & +4.09 & -1.44          &  +4.41
\end{tabular}}
\end{table}

\begin{figure}[t!]
    \centering
    \includegraphics[width=\linewidth, trim={0cm 0.1cm 0cm 0.2cm},clip]{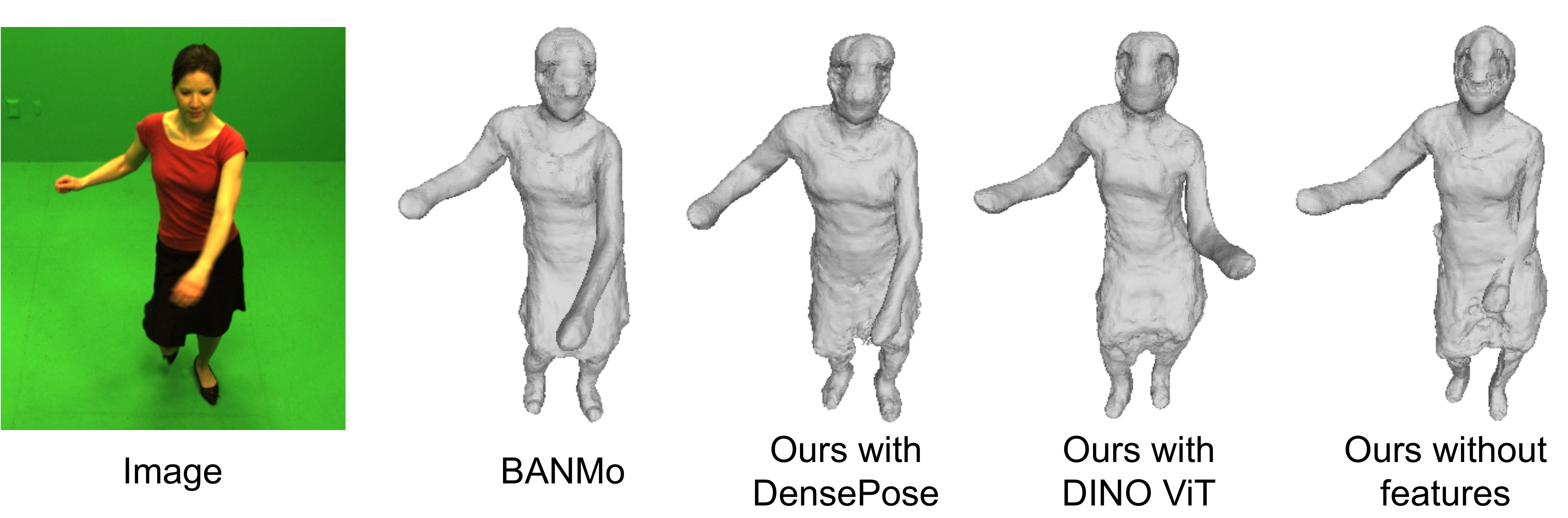}
    \caption{\textbf{Ablation on canonical features.} All methods recover high-fidelity surface, but only BANMo and our approach with DensePose supervision correctly predict the left arm's pose.}
    \label{fig:canonical_feature_ablation}
\end{figure}

\noindent\textbf{Canonical features}. DensePose CSE feature embeddings~\cite{neverova2020continuous} provide pre-trained 2D-3D correspondences for humans and are used by BANMo~\cite{banmo} on the AMA dataset. We also test our approach by replacing the DINO-ViT~\cite{DINO} features with the same DensePose CSE feature embeddings to supervise our canonical features. As shown in~\cref{fig:canonical_feature_ablation}, the arm is merged with the human body if we disable the canonical feature learning in our method. This is because the model cannot differentiate between the human body and the arm at this pose due to the lack of the object parts level visual cues. With DINO-ViT features, our method can separate the arm from the human body. However, it does not accurately predict the pose of the arm. The small gap between our approach with DINO-ViT and BANMo in~\cref{tab:reconstruction_comparision} mainly comes from these inaccurate pose predictions because the quality of the reconstructed mesh surface is similar, as shown in~\cref{fig:reconstruction_figs}.

\begin{figure}[t!]
    \centering
    \includegraphics[width=\linewidth, trim={0cm 0.1cm 0cm 0.2cm},clip]{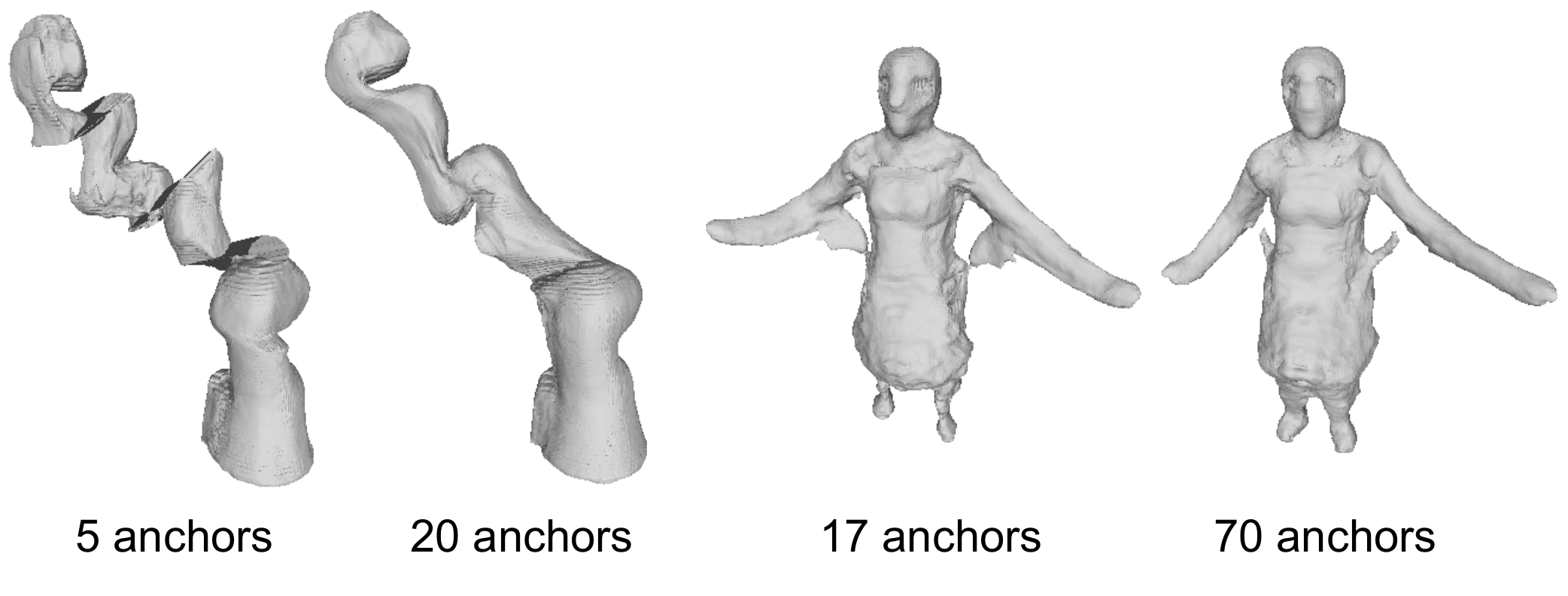}
    \caption{\textbf{Ablation on number of anchors.} We show the results of using half and double amounts of anchors on iiwa and AMA.}
    \label{fig:number_of_anchors_ablation}
    \vspace{-10pt}
\end{figure}

\noindent\textbf{Number of deformation anchors}. The number of deformation anchors is an object-specific hyper-parameter that plays a crucial role in modeling surface deformations. We show the results of using half and double amounts of anchors on AMA and iiwa datasets in~\cref{fig:number_of_anchors_ablation}. Having a small number of anchors limits our method's ability to model deformations at the robot arm's joints and the human's elbows, while having too many anchors makes the optimization process harder due to the over-parameterization of the deformations at the object's surface.

\section{Conclusion}
\label{sec:conclusion}
In this paper, we present a novel reconstruction pipeline that builds an animatable 3D model for any articulated objects from a collection of monocular videos. We leverage the foreground masks, optical flow, and pre-trained image features to iteratively optimize the kinematic chain along with the object's shape, appearance and deformation parameters. Our method can be generalized to any articulated object as it does not rely on category-specific priors. It allows users to render the object in arbitrary viewpoints and to perform pose manipulations in 3D directly while achieving state-of-the-art 3D surface reconstruction results. However, our kinematic chain does not check for unreachable states, such as chain collisions and foldings. We leave the task of learning physically plausible kinematic chains to future work. In addition, our deformation anchors are optimized based on the observed deformations from training videos. Therefore, our model does not generalize well to the novel poses that involve unseen object surface deformations.

\appendix
\section{Appendix Summary}
In Section~\ref{sec:method_details}, we present additional details on the method. Section~\ref{sec:implementation_details} provides additional details on the loss functions and optimization process. In Section~\ref{sec:additional_results}, we show more experimental results on each dataset. In Section~\ref{sec:anchors_asso}, we show examples of learned anchors and their corresponding associations. Section~\ref{sec:discussion_on_two_stage} provides additional discussions on the proposed two-stage optimization pipeline. Section~\ref{sec:limitations} and Section~\ref{sec:negative_impact} include the discussions on our method's limitations and potential negative impact. In Section~\ref{sec:failure_cases}, we show examples of failure cases. In Section~\ref{sec:iiwa_dataset}, we provide additional details on our iiwa dataset. Finally, we provide a list of the important variables used in the paper along with their state space and descriptions in Section~\ref{sec:notations}.

\section{Method Details}
\label{sec:method_details}

\subsection{Canonical representation}
\label{sec:additional_info_canonical}
We use the same canonical appearance, shape and feature representations as in BANMo~\cite{banmo}, where the color $\mathbf{c} \in \mathbb{R}^3$, the Signed Distance Function (SDF) value $\mathbf{v} \in \mathbb{R}$, and the canonical feature $\boldsymbol{\phi} \in \mathbb{R}^{16}$ of a point $\mathbf{x} \in \mathbb{R}^3$ in the canonical space are given by separate Multi-Layer Perception (MLP) Networks $\mathcal{F}_{\mathbf{C}}$, $\mathcal{F}_{\mathbf{S}}$, and $\mathcal{F}_{\boldsymbol{\phi}}$, respectively:
\begin{align}
    \mathbf{c} &= \mathbf{\mathcal{F}}_\mathbf{C}({\mathbf{x}}, {\mathbf{d}},\boldsymbol{\psi}_l) \notag\\
    \mathbf{v} &= \mathbf{\mathcal{F}}_{\mathbf{S}}({\mathbf{x}}) \notag\\
    \boldsymbol{\phi} &= \mathbf{\mathcal{F}}_{\boldsymbol{\phi}}({\mathbf{x}}),
\end{align}
where $\mathbf{d} \in \mathbb{R}^2$ is the viewing direction and $\boldsymbol{\psi}_l \in \mathbb{R}^{64}$ is a learnable latent code that captures the environment illumination conditions following~\cite{martin2021nerf}. Similar to ~\cite{yariv2021volume,wang2021neus}, we compute the density $\boldsymbol{\sigma} \in \mathbb{R}$ in the canonical space as:
\begin{equation}
    \boldsymbol{\sigma} = \Psi_{\beta}(\mathbf{v}),
\end{equation}
where $\Psi_{\beta}$ is cumulative of a Laplacian distribution with zero mean and learnable scale $\beta$, and $\mathbf{v}$ is the value of the SDF at the point $\mathbf{x}$.

To perform volumetric rendering for the pixel of interest $\mathbf{x}_I^t \in \mathbb{R}^2$ at time $t$, we first sample N points along the camera ray corresponding to the pixel in deformed space. As our implicit representations of the shape and appearance are defined in canonical space, we first transform the sampled points back to the canonical space~\cite{banmo, park2021nerfies} using the learned backward kinematics. Let the $i$-th point along the canonical space camera ray be $\mathbf{x}_i$, and its corresponding color and density queried from the MLP networks be $\mathbf{c}_i$ and $\boldsymbol{\sigma}_i$, we then follow the volumetric rendering process in NeRF~\cite{mildenhall2020nerf} to get the rendered color at the pixel of interest $\mathbf{c}_r$ as:
\begin{align}
    \mathbf{c}_r &= \sum_{i=1}^{N} T_i(1-\exp(-\boldsymbol{\sigma}_i\delta_i))\mathbf{c}_i \notag\\
    T_i&=\exp \bigg (-\sum_{j=1}^{i-1} \boldsymbol{\sigma}_j\delta_j \bigg ),
\end{align}
where $T_i$ is the accumulated transmittance between the camera center to the $i$-th sampled point, $\delta_i$ is the interval between consecutive sampled points.

\subsection{Recovering proper kinematic chain}
In this section, we provide additional details on how to recover a proper and connected kinematic chain after all the joints in the canonical space are transformed into the deformed space using unconstrained transformations. 

Given the canonical kinematic chain as $\mathbf{P}=\left\{\mathbf{p}_{i} \mid i=1, \ldots, {n}_{j}\right\}$, and the initial optimization stage's learned unconstrained anchor transformations $\mathbf{\hat T}^t$ from the canonical space to the deformed space, we can then apply $\mathbf{\hat T}^t$ to transform the canonical kinematic chain joints $\mathbf{P}$ to the deformed space as:
\begin{equation}
\label{eq:forward_kinematic_chain_appendix}
{\mathbf{\hat P}}^{t} = {\mathbf{C}}^{t} \sum_{i=1}^{{n}_{a}} ({\mathbf{w}}_{{a}_{i}} {\bf \hat T}^t_{i}) \mathbf{P},
\end{equation}
where $\mathbf{\hat P}^t=\left\{\mathbf{\hat p}_{i} \mid i=1, \ldots, {n}_{p}\right\}$ is the set of unconstrained kinematic chain joints in the deformed space at time $t$. These unconstrained joints cannot form a proper kinematic chain that preserves the hierarchical order of connections and consistent kinematic chain length, because the unconstrained transformations $\mathbf{\hat T}^t$ are learned without any knowledge of the kinematic chain. To recover the proper kinematic chain, we follow~\cref{alg:recovering_alg} to obtain the revised location of the joints in hierarchical order denoted by $\mathbf{\tilde P}=\left\{\mathbf{\tilde p}_{i} \mid i=1, \ldots, {n}_{j}\right\}$. Given the joints' revised locations, we can infer the anchors' revised transformations $\mathbf{\tilde T}^t$ by enforcing pre-defined associations between anchors and the kinematic chain links in the deformed space.

\begin{algorithm}
\caption{Recovering proper kinematic chain}\label{alg:recovering_alg}
\begin{algorithmic}
\For {$j = {1}:{n}_j$ in \textbf{hierarchical order}}
\For{$\mathbf{p}_k \in {\mathbf{Children}({\mathbf{p}_j})}$}
    \State $\boldsymbol{\mu} \gets \frac{\left \| \mathbf{p}_{k} - \mathbf{p}_{j}\right \|_2}{\left \| \mathbf{\hat p}_{k} - \mathbf{\hat p}_j \right \|_2}$ 
    \State $\mathbf{\tilde p}_k = \mathbf{\tilde p}_j + \boldsymbol{\mu}(\mathbf{\hat p}_{k} - \mathbf{\hat p}_j)$
    \State ${\mathbf{t}_k} \gets \mathbf{\tilde p}_{k} - \mathbf{\hat p}_{k}$ 
    \For{$\mathbf{p}_d \in {\mathbf{Descendants}({\mathbf{p}_k})}$}
        \State $\mathbf{\hat p}_d \gets \mathbf{\hat p}_d + \mathbf{t}_k$
    \EndFor
\EndFor
\State {$j \gets j+1$}
\EndFor
\end{algorithmic}
\end{algorithm}

\subsection{Additive residuals to kinematic chain}
We introduce a learnable additive residual term for each kinematic chain link during the kinematic chain optimization stage. Given the canonical kinematic chain as $\mathbf{P}=\left\{\mathbf{p}_{i} \mid i=1, \ldots, {n}_{j}\right\}$ and its corresponding links as  $\mathbf{L}=\left\{\mathbf{\boldsymbol {\ell}}_{jk}=(\mathbf{p}_{j},\mathbf{p}_{k})\right\}^{({n}_{j}-1)}$, let $\mathbf{R}=\left\{\mathbf{r}_{i} \mid i=1, \ldots, ({n}_{j}-1) \right\}$ be the set of learnable residuals. We clip the residuals to avoid drastic changes on the kinematic chain initialization for stable training as:
\begin{equation}
    \mathbf{\tilde R} = \boldsymbol{\gamma}\tanh(\mathbf{R}),
    \label{eq:clip_residual}
\end{equation}
where $\mathbf{\tilde R}=\left\{\mathbf{\tilde r}_{i} \mid i=1, \ldots, ({n}_{j}-1) \right\}$ are the set of clipped residuals and $\boldsymbol{\gamma}$ is a hyper-parameter that ensures the change of each kinematic chain link's length is within $\left( -\boldsymbol{\gamma}, \boldsymbol{\gamma} \right )$. 

We update the canonical kinematic chain with the latest residuals $\mathbf{\tilde R}$ in each iteration to obtain the updated kinematic chain $\mathbf{\tilde P}=\left\{\mathbf{\tilde p}_{i} \mid i=1, \ldots, {n}_{j}\right\}$ following~\cref{alg:residual_update}. Specifically, we start from the root joint of the kinematic chain and update each link's length by changing the corresponding joint's position in a hierarchical manner to ensure that we do not break the kinematic chain. After the residual update, we compute the new set of association parameters for each anchor with respect to the updated kinematic chain for kinematic chain driven deformations.

\begin{algorithm}
\caption{Kinematic chain update with residuals}
\label{alg:residual_update}
\begin{algorithmic}
\For {$j = {1}:{n}_j$ in \textbf{hierarchical order}}
\For{$\mathbf{p}_k \in {\mathbf{Children}({\mathbf{p}_j})}$}
    \State $\boldsymbol{\mu} \gets \frac{\left \| \mathbf{p}_{k} - \mathbf{p}_{j}\right \|_2 + \mathbf{\tilde r}_k}{\left \| \mathbf{p}_{k} - \mathbf{p}_j \right \|_2}$ 
    \State $\mathbf{\tilde p}_k = \mathbf{\tilde p}_j + \boldsymbol{\mu}(\mathbf{p}_{k} - \mathbf{p}_j)$
    \State ${\mathbf{t}_k} \gets \mathbf{\tilde p}_{k} - \mathbf{p}_{k}$ 
    \For{$\mathbf{p}_d \in {\mathbf{Descendants}({\mathbf{p}_k})}$}
        \State $\mathbf{p}_d \gets \mathbf{p}_d + \mathbf{t}_k$
    \EndFor
\EndFor
\State {$j \gets j+1$}
\EndFor
\end{algorithmic}
\end{algorithm}

\section{Additional Implementation Details}
\label{sec:implementation_details}
In this section, we provide additional details on the loss functions and optimizations.

\subsection{Loss functions}
Similar to~\cite{banmo}, we impose reconstruction losses on the 2D observations, including 2D images, foreground masks, and optical flow. We follow the volumetric rendering process described in~\ref{sec:additional_info_canonical} to obtain predicted color and foreground mask values. To render optical flow, we follow~\cite{banmo} to transform canonical space points to consecutive frames' deformed spaces, and use the camera projection model to find their corresponding positions on the 2D image. We then compute their difference as the optical flow at the point of interest. The reconstruction losses are formulated in the same way as in~\cite{mildenhall2020nerf,banmo,yang2021viser,yang2021lasr,yariv2020multiview}, where we compute the difference between the rendered and the actual observations:
\begin{equation}
    \label{eq:reconstruction_losses_appendix}
    \mathcal{L}_{\textrm{recon}} = \sum_{\mathbf{x}_I} \left \| \mathbf{o}_{r} - \mathbf{o}_{gt} \right \|^2,
\end{equation}
where $\mathbf{o}_{r}$ and $\mathbf{o}_{gt}$ are the pairs of rendered and ground-truth 2D observations (2D images, foreground masks, optical flow) at pixels of interest $\mathbf{x}_I^t \in \mathbb{R}^2$ at time $t$.

We apply soft argmax descriptor matching~\cite{kendall2017end,luvizon2019human, yang2021viser, banmo} at the pixel of interest $\mathbf{x}_I^t$, to find the most probable corresponding surface point $\mathbf{x}_{m} \in \mathbb{R}^3$ in the canonical space by matching learned canonical feature embeddings at the object's surface with the pre-trained DINO ViT~\cite{DINO} features. Let $\mathbf{x}_c \in \mathbb{R}^3$ be the point that corresponds to the pixel of interest $\mathbf{x}_I^t$ based on backward kinematics. To supervise the canonical feature embedding, we follow~\cite{banmo} to minimize the difference between these two points' positions as:
\begin{equation}
    \label{eq:matching_loss}
    \mathcal{L}_{\textrm{3d-match}} = \sum_{\mathbf{x}_I} \left \| \mathbf{x}_{m} - \mathbf{x}_{c} \right \|^2_2.
\end{equation}
Let $\pi^t$ be the camera project model at time $t$, we transform $\mathbf{x}_m$ from the canonical space to the deformed space using learned forward kinematics, and use $\pi^t$ to project it to image coordinates. We follow~\cite{banmo,kulkarni2020articulation,yang2021viser} to enforce the consistency at the image coordinate level by minimizing the difference between the obtained point and $\mathbf{x}_I^t$:
\begin{equation}
    \label{eq:2d_consistency}
    \mathcal{L}_{\textrm{2d-match}} = \sum_{\mathbf{x}_I} \left \|  \pi^t \bigg ( {\mathbf{C}}^{t} \sum_{i=1}^{{n}_{a}} ({\mathbf{w}}_{{a}_{i}} {\bf T}^t_{i}) \mathbf{x}_{m} \bigg ) - \mathbf{x}_I^t \right \|^2_2.
\end{equation}
In addition, we follow~\cite{banmo,li2021neural} to encourage consistency in terms of learned forward kinematics and backward kinematics. Specifically, we transform each point $\mathbf{x}_i^t$ along the sampled camera ray in the deformed space to the canonical space using backward kinematics, and then transform each point back to the deformed space. We encourage the resulting point $\mathbf{x}_i^{t'}$to be at the same position as $\mathbf{x}_i^t$:
\begin{equation}
    \label{eq:3d_consistency}
    \mathcal{L}_{\textrm{transform}} = \sum_{i} T_i \left \| \mathbf{x}_{i}^{t'} - \mathbf{x}_{i}^{t} \right \|^2_2,
\end{equation}
where $\mathbf{x}_{i}^{t}$ is the original sampled point along the camera ray, $\mathbf{x}_{i}^{t'}$ is the sampled point that undergoes a backward and a forward transform, and $T_i$ the accumulated transmittance between the camera center to the $i$-th sampled point, which we use as the weight for individual sampled points.

During the kinematic chain optimization stage, we also introduce regularization on the anchors and learned transformations by minimizing the differences between the revised anchors based on the kinematic chain and the unconstrained anchors from learned unconstrained anchor transformations $\mathbf{\hat T}_i^t$ predicted by $\mathbf{\mathcal{F}_{A}}$:
\begin{equation}
    \label{eq:anchors_loss}
    \mathcal{L}_{\textrm{anchors}} = \sum_{i=1}^{n_a} \left \| \mathbf{\tilde a}_i - \mathbf{\hat T}_i^t \mathbf{a}_i \right \|^2_2,
\end{equation}
where $\mathbf{\tilde a}_i$ is the revised anchor based on kinematic chain constraints and $\mathbf{\hat T}_i^t$ is the unconstrained rigid transformation from the canonical space to the deformed space given by the MLP network $\mathbf{\mathcal{F}_{A}}$ for the canonical space anchor $\mathbf{a}_i$.

\subsection{Optimizations}
Our model is trained using the Adam~\cite{kingma2014adam} optimizer with weight decay on a single RTX 3090 GPU. In the initial optimization stage, we use an initial learning rate of $0.0005$ and update it using one-cycle policy~\cite{smith2018disciplined} and cosine annealing~\cite{loshchilov2016sgdr}. Inspired by~\cite{banmo}, we introduce a separate MLP network after the initial optimization stage that takes in a 3D point, and outputs a skinning weight residual with respect to each anchor that adds to the Euclidean distance based skinning weights before normalization, to obtain smoother deformed surface. All the MLP networks and learnable parameters are optimized jointly during training. For the iiwa dataset, we do not update the kinematic chain link lengths with additive residuals because iiwa is considered a rigid robot arm without deformable surface, and its kinematic chain initialization is generally a single chain that already captures its articulation modes.

\section{Additional Results}
\label{sec:additional_results}

\subsection{More qualitative results}
\begin{figure}[t!]
    \centering
    \includegraphics[width=\linewidth, trim={0cm 0.1cm 0cm 0.2cm},clip]{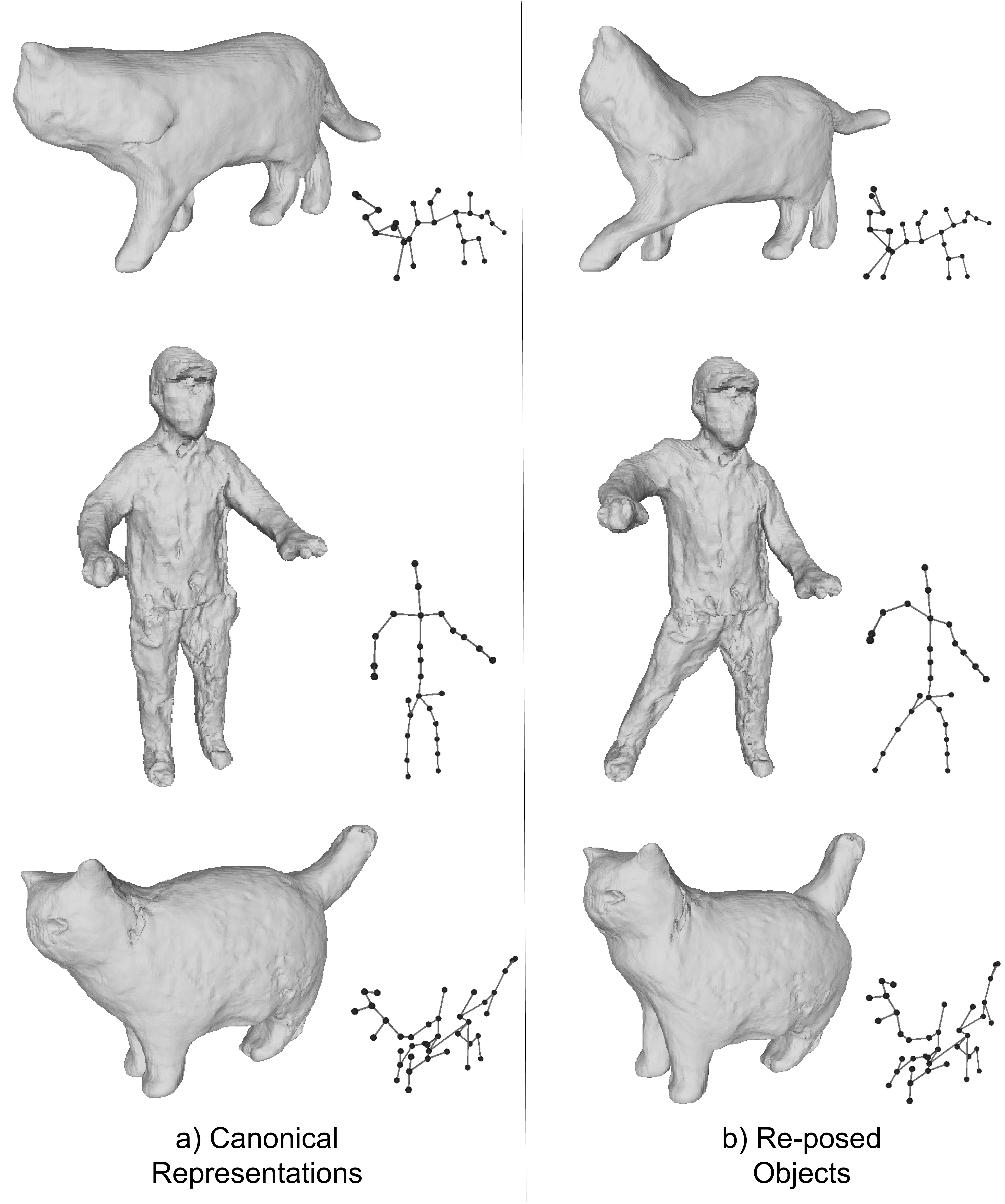}
    \caption{\textbf{Pose manipulation examples}. In a) we show the learned canonical representations of the objects for in-the-wild datasets. We perform pose manipulations using optimized kinematic chain and show the reposed kinematic chain and meshes in b).} 
    \label{fig:in_the_wild_repose}
\end{figure}

\begin{figure}[t!]
    \centering
    \includegraphics[width=\linewidth, trim={0cm 0.1cm 0cm 0.2cm},clip]{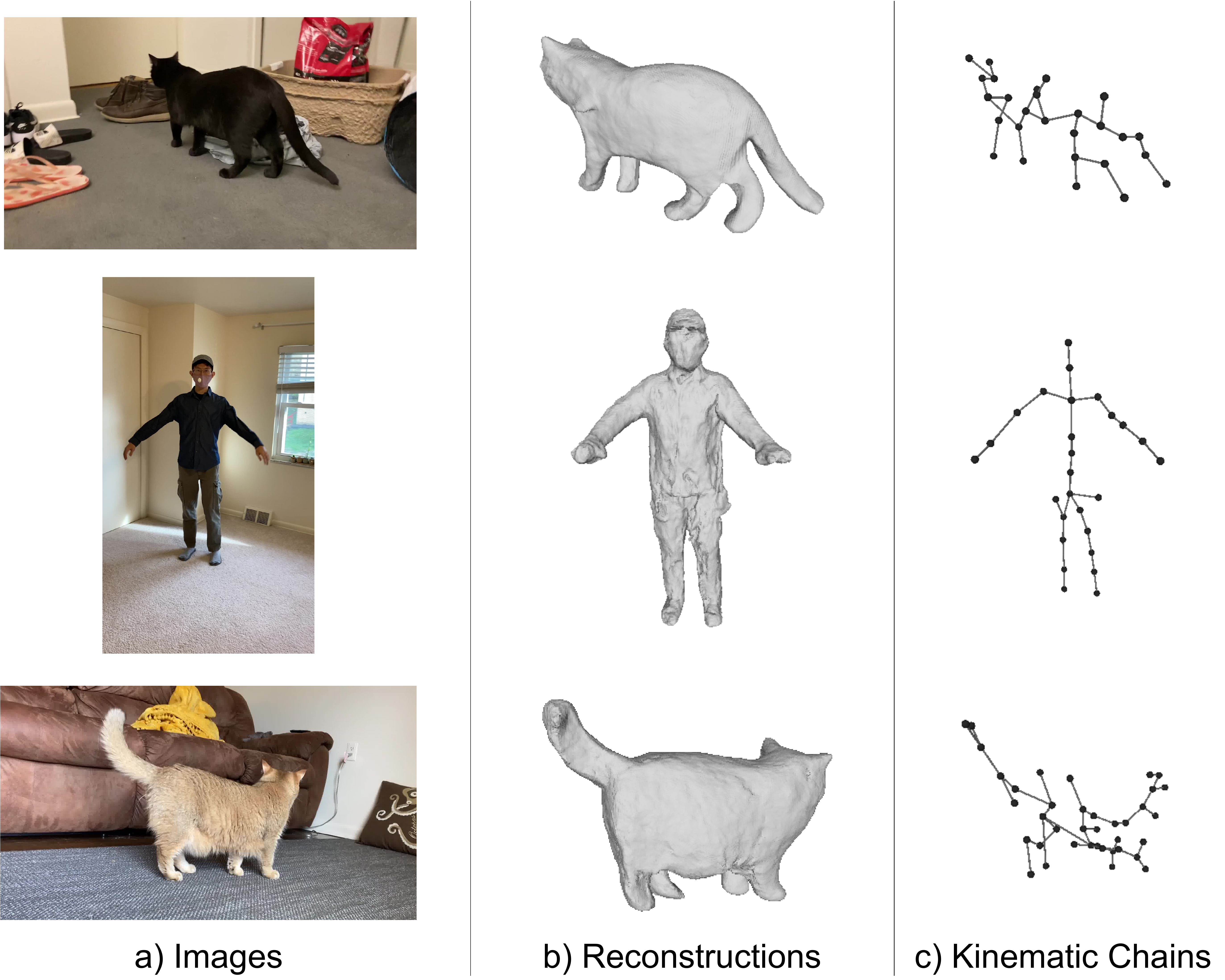}
    \caption{\textbf{3D surface reconstruction results.} In a) we show some RGB images from the in-the-wild datasets. In b) we show the reconstructed 3D meshes and in c) we show the corresponding kinematic chains for the same frame.} 
    \label{fig:in_the_wild_recon}
    \vspace{-10pt}
\end{figure}

We test on in-the-wild monocular captures of various deformable and articulated objects~\cite{banmo}. We use off-the-shelf pre-trained models to obtain foreground masks~\cite{kirillov2020pointrend}, initial root pose estimation~\cite{banmo} and optical flow~\cite{yang2019vcn} for each frame of the in-the-wild dataset videos. We show manual pose manipulation results in~\cref{fig:in_the_wild_repose}, and 3D surface reconstruction results in~\cref{fig:in_the_wild_recon} for in-the-wild datasets. 

We also include video examples of manual pose manipulations by re-posing the kinematic chain, and the 3D reconstruction results for poses in training videos for each dataset. Please refer to the videos in the supplementary material for better visualization.

\subsection{More quantitative results}
We report quantitative 3D surface reconstruction results for iiwa~(\cref{tab:iiwa_all_results}), Eagle~(\cref{tab:eagle_all_results}), AMA-swing~(\cref{tab:ama_swing_all_results}), and AMA-samba~(\cref{tab:ama_samba_all_results}) evaluated in 3D Chamfer Distances in centimeters and F-scores at distance thresholds of 1$\%$, 2$\%$, and 5$\%$. Note that for the AMA dataset, we train each model using all the sequences (both swing and samba), and report results on swing and samba separately.

\begin{table}
\centering
\caption{3D surface reconstruction results on \textbf{iiwa} dataset evaluated in 3D Chamfer Distances ($\downarrow$) and F-scores ($\uparrow$) at distance thresholds of 1$\%$, 2$\%$, and 5$\%$ averaged across all frames.}
\resizebox{\columnwidth}{!}{
\begin{tabular}{c|cccc}
\label{tab:iiwa_all_results}
 Method  &  CD            & F@1\%  &  F@2\% &  F@5\% \\
 \midrule
 5 anchors & 7.40 & 26.76 & 51.16 & 79.84\\
 10 anchors & 5.52 & 28.69 & 56.53 & 85.61\\
 20 anchors & 5.95 & 28.56 & 55.17 & 83.51\\
\midrule
 with DensePose feat. & 5.64 & 28.85 & 55.67 & 85.79 \\
 w/o feat. & 7.26 & 25.16 & 47.33 & 73.82 \\
\midrule
 w/o kine. chain opt. & 5.69 & 26.90 & 52.44 & 81.81
\end{tabular}
}
\end{table}

\begin{table}
\centering
\caption{3D surface reconstruction results on \textbf{Eagle} dataset evaluated in 3D Chamfer Distances ($\downarrow$) and F-scores ($\uparrow$) at distance thresholds of 1\%, 2\%, and 5\% averaged across all frames.}
\resizebox{\columnwidth}{!}{
\begin{tabular}{c|cccc}
\label{tab:eagle_all_results}
 Method  &  CD            & F@1\%  &  F@2\% &  F@5\% \\
 \midrule
 12 anchors & 4.51 & 43.20 & 81.86 & 98.88\\
 25 anchors & 4.21 & 43.41 & 83.38 & 99.22\\
 50 anchors & 4.38 & 42.21 & 82.52 & 98.91\\
\midrule
 with DensePose feat. & 4.31 & 43.07 & 82.45 & 99.11 \\
 w/o feat. & 4.31 & 43.51 & 82.81 & 98.98 \\
\midrule
 w/o kine. chain opt. & 7.44 & 25.30 & 57.87 & 94.40
\end{tabular}
}
\end{table}

\begin{table}
\centering
\caption{3D surface reconstruction results on \textbf{AMA-swing} evaluated in 3D Chamfer Distances ($\downarrow$) and F-scores ($\uparrow$) at distance thresholds of 1\%, 2\%, and 5\% averaged across all frames.}
\resizebox{\columnwidth}{!}{
\begin{tabular}{c|cccc}
\label{tab:ama_swing_all_results}
 Method  &  CD            & F@1\%  &  F@2\% &  F@5\% \\
 \midrule
 17 anchors & 12.41 & 24.37 & 45.69 & 78.10\\
 35 anchors & 9.69 & 29.17 & 53.29 & 85.20\\
 70 anchors & 12.49 & 25.62 & 46.72 & 77.45\\
\midrule
 with DensePose feat. & 8.96 & 33.68 & 58.46 & 86.49 \\
 w/o feat. & 9.88 & 27.60 & 52.15 & 84.42 \\
\midrule
 w/o kine. chain opt. & 11.13 & 26.54 & 48.88 & 80.69
\end{tabular}
}
\end{table}

\begin{table}
\centering
\caption{3D surface reconstruction results on \textbf{AMA-samba} evaluated in 3D Chamfer Distances ($\downarrow$) and F-scores ($\uparrow$) at distance thresholds of 1\%, 2\%, and 5\% averaged across all frames.}
\resizebox{\columnwidth}{!}{
\begin{tabular}{c|cccc}
\label{tab:ama_samba_all_results}
 Method  &  CD            & F@1\%  &  F@2\% &  F@5\% \\
 \midrule
 17 anchors & 11.76 & 25.51 & 49.09 & 81.52\\
 35 anchors & 9.22 & 29.81 & 54.20 & 86.91\\
 70 anchors & 11.72 & 25.74 & 48.72 & 80.69\\
\midrule
 with DensePose feat. & 8.34 & 33.89 & 60.16 & 88.66 \\
 w/o feat. & 9.24 & 27.06 & 52.88 & 87.37 \\
\midrule
 w/o kine. chain opt. & 9.80 & 31.86 & 55.34 & 84.69
\end{tabular}
}
\end{table}

\section{Anchors and Associations}
\label{sec:anchors_asso}

\begin{figure}[t!]
    \centering
    \includegraphics[width=\linewidth, trim={0cm 0.1cm 0cm 0.2cm},clip]{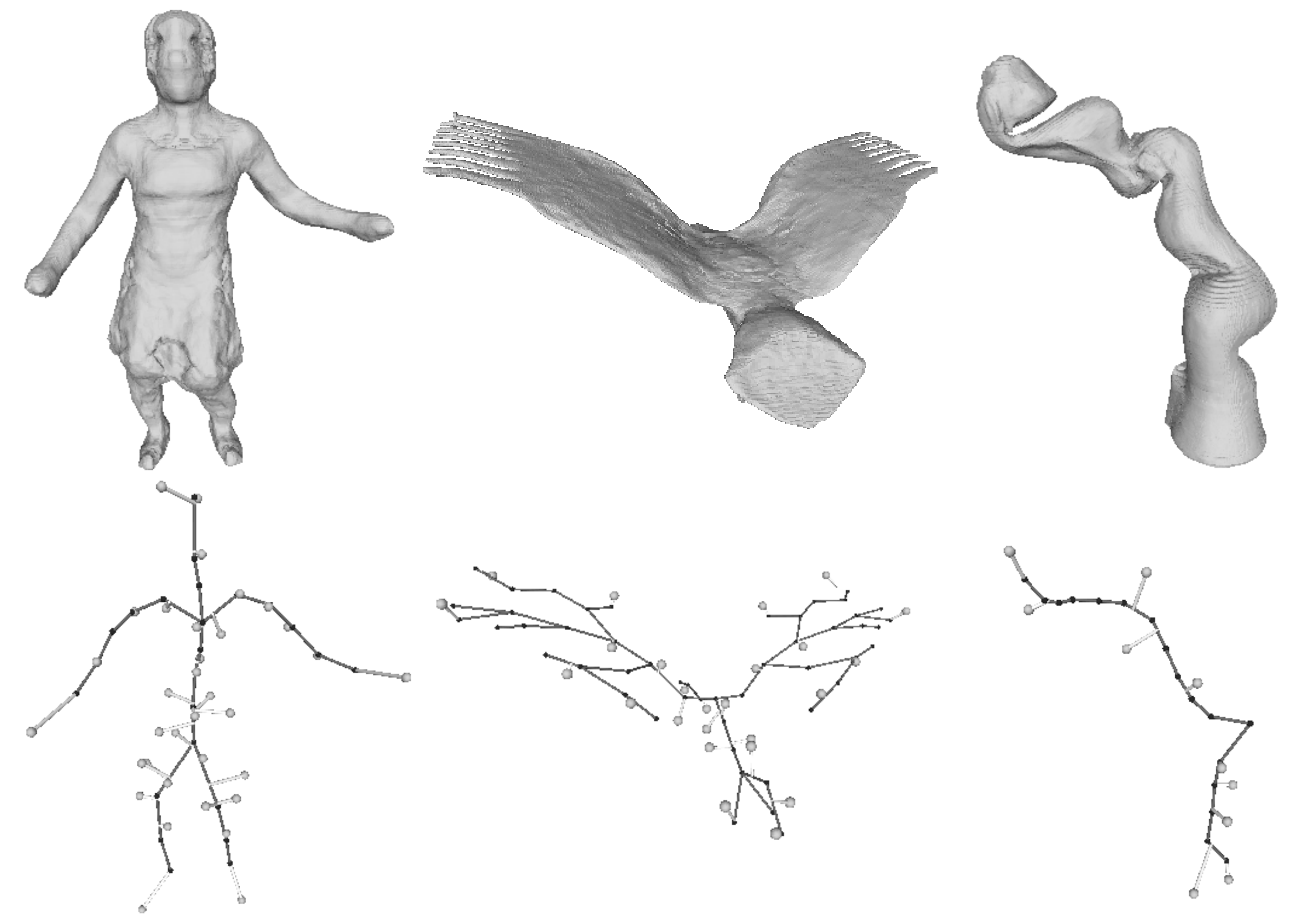}
    \caption{\textbf{Anchors and associations.} Anchors (white dots around the kinematic chain) and their corresponding associations (line segments between anchors and kinematic chain links) for the synthetic and AMA datasets.} 
    \label{fig:anchors_fig}
    \vspace{-10pt}
\end{figure}

The anchors are optimized to model the modes of the deformations, and the skinning weights are optimized w.r.t. anchors. By enforcing deterministic associations between the anchors and the kinematic chain, the transformations of anchors can be directly inferred from user-defined kinematic chain transformations. Therefore, we can make good use of optimized anchors and their corresponding skinning weights, instead of optimizing the skinning weights w.r.t. kinematic chain joints from scratch. In~\cref{fig:anchors_fig}, we show the learned deformation anchors and their associations to the kinematic chain links for each dataset. The anchors move along with the kinematic chain links to keep the association parameters constant at all times to enable re-posings directly driven by the kinematic chain. 

Note that our proposed kinematic chain driven deformations formulation is capable of modeling twists of a kinematic chain link about its axial direction even though our kinematic chain is defined as connected line segments. An additional rotation matrix that represents the rotation about the axial axis of the link is pre-multiplied to rotate the anchor around its associated link for twist effects.

\section{Necessity of Two-Stage Optimization}
\label{sec:discussion_on_two_stage}
Since we do not have access to any shape prior or template, building both object's shape and kinematic chain from scratch simultaneously is a highly ill-conditioned optimization problem given only a collection of monocular videos. Our proposed two-stage optimization simplifies this problem by taking advantage of the techniques in the existing template-free surface reconstruction method~\cite{banmo} to extract an initial estimate of the 3D shape, which allows us to use RigNet~\cite{xu2020rignet} to extract an initial estimate of the kinematic chain. The two-stage scheme significantly reduces the search space for the underlying rigid structure of the object and makes the optimization much more stable. 

As RigNet~\cite{xu2020rignet} operates on 3D mesh, it's possible to directly apply RigNet to the initial optimization stage results to obtain an animtable model driven by RigNet's predicted skinning weights. However, it would yield three problems. The first problem is that the skinning weights predicted by RigNet are based on a static mesh without any knowledge from the videos (dynamic scenes). These weights are inaccurate and not optimized for the deformation modes that occurred in the videos. The second problem is that RigNet suffers from significantly large memory consumption and runtime. The author of RigNet recommends users to only predict skinning weights for kinematic chain joints on low-resolution mesh ($\leq$ 5,000 vertices) on its code repository~\cite{xu2020rignet} to prevent running out of hardware's memory and extremely long runtime. Our approach does not have this bottleneck as we do not rely on the skinning weights predicted by RigNet. We build the associations between anchors and the kinematic chain to animate object meshes regardless of the resolution. The third problem is that directly using the RigNet initialization combined with the anchors would lead to poor results. As reported in the Tables in the main paper and the supplementary material, we achieve consistent improvements on all datasets with our proposed kinematic chain optimization stage.

\begin{figure}[t!]
    \centering
    \includegraphics[width=\linewidth, trim={0cm 0.1cm 0cm 0.2cm},clip]{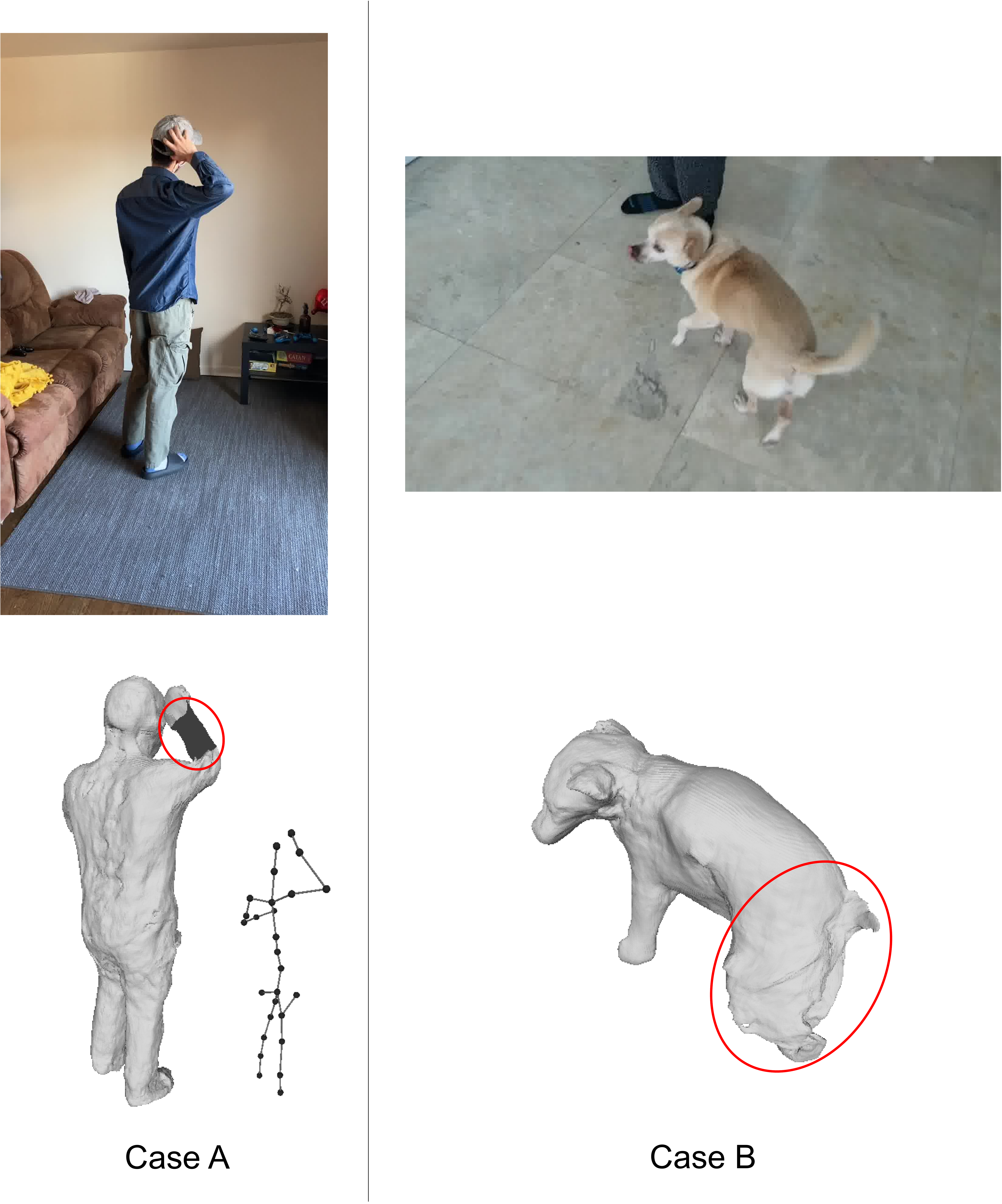}
    \caption{\textbf{Failure case examples.} In Case A, the mesh of the person's right arm collapses despite the kinematic chain being able to represent the correct pose. In Case B, the dog's tail and back legs are missing due to bad reconstruction.} 
    \label{fig:failure_cases}
    \vspace{-10pt}
\end{figure}

\section{Limitations}
\label{sec:limitations}

\noindent\textbf{Input dependencies}. In our pipeline, we leverage an off-the-shelf model~\cite{yang2019vcn} to extract optical flow from monocular videos, and use ground-truth foreground masks from the datasets for optimizations. For the Eagle and iiwa datasets, we use the ground-truth root poses, while for the AMA dataset, we leverage a pre-trained PoseNet~\cite{banmo} for root pose initializations and jointly optimize them during training. Therefore, our pipeline's performance is affected by the quality of these inputs, and it requires a generic root pose estimator to enable our pipeline to work on any object of interest because PoseNet~\cite{banmo} is only suitable for humans and quadruped animals.

\noindent\textbf{Kinematic chain initialization}. We use a category-agnostic skeleton estimator, RigNet~\cite{xu2020rignet}, on the initial 3D mesh estimate of the object to obtain kinematic chain initialization. However, the underlying structures and articulation modes vary across different object categories, requiring users to tune a RigNet's test-time hyper-parameter that controls how dense the joints are distributed within the object's mesh. We encourage the users to run RigNet multiple times on the same initial estimate of the mesh with different test-time hyper-parameters, and select the kinematic chain that agrees with the object's underlying structure and articulation modes the most before optimizing it.

\section{Potential Negative Impact}
\label{sec:negative_impact}
As our approach effectively builds a 3D animatable model of an object using monocular videos, it can be potentially used for malicious purposes, such as generating fake videos or other formats of illegal content. Since monocular videos are often easy to acquire in the real world, it is important to ensure our method is used with prior consent.

\section{Failure Cases}
\label{sec:failure_cases}

We show some example failure cases in~\cref{fig:failure_cases}. In Case A, the mesh of the person's right arm collapses at the pose shown in the figure despite the correct kinematic chain pose. This is because the deformation anchors are not optimized well to handle the desired pose. In the cases of training with fast-moving objects and large self-occlusions in the videos, the reconstruction results sometimes suffer from significant degradation, as shown in Case B. The dog's tail is moving very fast in the training video and often causes self-occlusions at the back of the dog, making the overall optimization much less stable and sometimes converge to poor results. In such cases, we recommend tuning the hyper-parameters in training or gathering more videos with 360 degrees captures of the object to make the training more stable. In general, objects that are moving slowly with 360 degrees captures are usually easier to train.

\begin{figure}[t!]
    \centering
    \includegraphics[width=\linewidth, trim={0cm 0.1cm 0cm 0.2cm},clip]{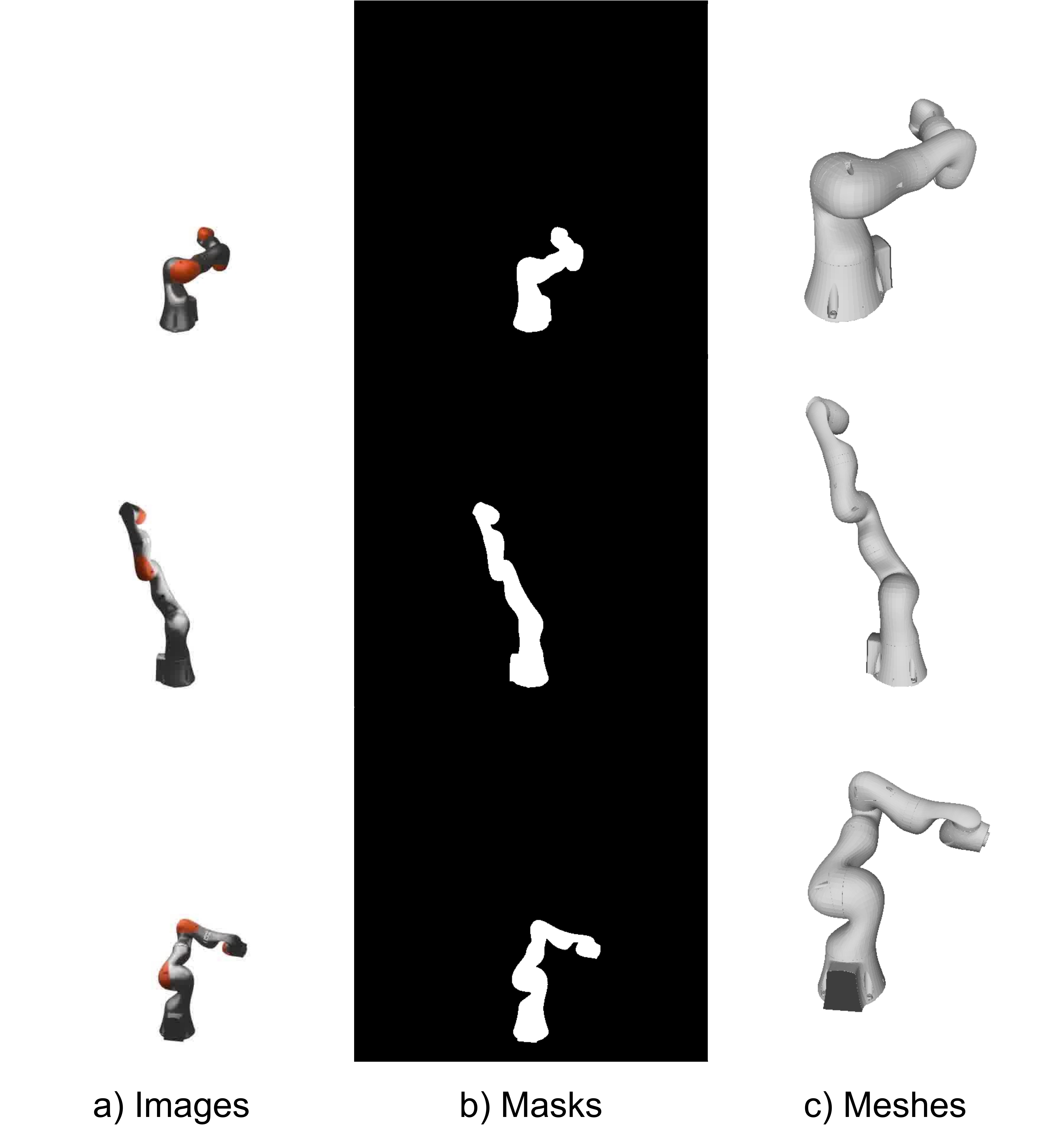}
    \caption{\textbf{Examples of our iiwa dataset.} We show examples of a) RGB images, b) corresponding foreground masks, and c) corresponding ground truth meshes for three randomly picked frames in the dataset. Note that the meshes are re-scaled for better visualization in the figure.} 
    \label{fig:iiwa_dataset}
    \vspace{-10pt}
\end{figure}

\section{Additional Details on iiwa Dataset}
\label{sec:iiwa_dataset}

We created a new dataset that contains four video sequences of an animated KUKA LBR iiwa robot arm. The synthetic model of the robot arm is obtained from BlenderKit. We animate the robot arm in Blender and render the videos of the robot arm's motion. Each video contains 400 frames in total. Note that for each frame in the video sequences, the camera pose, foreground mask, and 3D mesh of the robot arm are also included in the dataset for quantitative evaluation purposes. We show examples of our dataset in~\cref{fig:iiwa_dataset}.

\section{Notations}
\label{sec:notations}
In~\cref{tab:variable_list}, we list the important variables used in this paper, along with their state space and descriptions.

\begin{table*}
\caption{\textbf{Notations.} A list of the important variables used in the paper.}
\centering
\makebox[\textwidth]{
\begin{tabular}{lll}
\toprule
 \textbf{Symbol}  & \textbf{State space} & \textbf{Description} \\
 \midrule
  $\mathbf{\mathcal{F}}_\mathbf{C}$    & MLP       & MLP for color in canonical space\\
 $\mathbf{\mathcal{F}}_\mathbf{S}$    & MLP       & MLP for SDF in canonical space\\
 $\mathbf{\mathcal{F}}_{\boldsymbol{\phi}}$    & MLP       & MLP for canonical feature in canonical space\\
 $\mathbf{\mathcal{F}}_\mathbf{A}$    & MLP       & MLP for unconstrained transformations of anchors\\
 $\mathbf{p}_i$    & $\mathbb{R}^3$       & $i$-th kinematic chain joint in canonical space\\
$\mathbf{\hat p}_i$    & $\mathbb{R}^3$       & $i$-th unconstrained kinematic chain joint\\
$\mathbf{\tilde p}_i$    & $\mathbb{R}^3$       & $i$-th revised kinematic chain joint \\
$\mathbf{a}_{i}$    & $\mathbb{R}^3$       & $i$-th deformation anchor in canonical space\\
$\mathbf{\ell}_{jk}$    & link       & Kinematic chain link between $\mathbf{p}_j$ and $\mathbf{p}_k$\\
$\mathbf{C}^t$ & $SE(3)$ & Object root pose w.r.t the canonical root pose at time $t$\\
$\mathbf{T}_i^t$ & $SE(3)$ & Transformation of anchor $\mathbf{a}_i$ at time $t$\\
$\mathbf{\hat T}_i^t$ & $SE(3)$ & Unconstrained transformation of anchor $\mathbf{a}_i$  at time $t$\\
$\mathbf{\tilde T}_i^t$ & $SE(3)$ & Revised transformation of anchor $\mathbf{a}_i$ at time $t$\\
$\mathbf{H}_i^t$ & $SE(3)$ &  Forward kinematics of kinematic chain joint $\mathbf{p}_i$ at time $t$\\
$\mathbf{w}_{a_i}$ & $\mathbb{R}$ & Forward skinning weight w.r.t anchor $\mathbf{a}_i$\\
$\mathbf{w}_{a_i}^t$ & $\mathbb{R}$ & Backward skinning weight w.r.t anchor $\mathbf{a}_i$ at time $t$\\
$\mathbf{R}$ & $\mathbb{R}^{n_j-1}$ & Unclipped additive residuals for kinematic chain\\
$\mathbf{\tilde R}$ & $\mathbb{R}^{n_j-1}$ & Clipped additive residuals for kinematic chain\\
$\mathbf{x}_I^t$ & $\mathbb{R}^2$ & Pixel of interest at time $t$\\
$\boldsymbol\psi_a^t$ & $\mathbb{R}^{128}$ & Learnable latent code for anchor transformations at time $t$\\
$\boldsymbol\psi_l$ & $\mathbb{R}^{64}$ & Learnable latent code illumination conditions\\
$\pi^t$ & $\mathbb{R}^{3 \times 4}$  & Camera projection model at time $t$\\
 \bottomrule
\label{tab:variable_list}
\end{tabular}
}
\end{table*}
\clearpage

\clearpage
{\small
\bibliographystyle{ieee_fullname}
\bibliography{egbib}
}

\end{document}